\documentclass[sigconf]{acmart}

\usepackage{algorithm}
\usepackage{algorithmic}
\usepackage{multirow}
\usepackage{makecell}
\usepackage{tabularx}
\usepackage{bbding}
\usepackage{tabularray}
\usepackage{balance}

\newcommand{\eg}{\textit{e.g.}}
\AtBeginDocument{%
  }

\copyrightyear{2025} 
\acmYear{2025} 
\setcopyright{acmlicensed}\acmConference[MM '25]{Proceedings of the 33rd ACM International Conference on Multimedia}{October 27--31, 2025}{Dublin, Ireland}
\acmBooktitle{Proceedings of the 33rd ACM International Conference on Multimedia (MM '25), October 27--31, 2025, Dublin, Ireland}
\acmDOI{10.1145/3746027.3758196}
\acmISBN{979-8-4007-2035-2/2025/10}

\begin{document}

\title{WeatherBench: A Real-World Benchmark Dataset for All-in-One Adverse Weather Image Restoration}

\author{Qiyuan Guan}
\authornote{Both authors contributed equally to this research.}
\orcid{0009-0001-8557-1560}
\affiliation{%
  \institution{Dalian Polytechnic University}
  \city{Dalian}
  \country{China}
}
\email{csguanqiyuan@163.com}

\author{Qianfeng Yang}
\authornotemark[1]
\affiliation{%
  \institution{Dalian Polytechnic University}
  \city{Dalian}
  \country{China}
  }
\email{csqianfengyang@163.com}

\author{Xiang Chen}
\authornotemark[1]
\affiliation{%
  \institution{Nanjing University of Science and Technology}
  \city{Nanjing}
  \country{China}
  }
\email{chenxiang@njust.edu.cn}

\author{Tianyu Song}
\affiliation{%
  \institution{Dalian Martime University}
  \city{Dalian}
  \country{China}
  }
\email{songtienyu@163.com}

\author{Guiyue Jin}
\authornote{Corresponding author.}
\affiliation{%
  \institution{Dalian Polytechnic University}
  \city{Dalian}
  \country{China}
  }
\email{guiyue.jin@dlpu.edu.cn}

\author{Jiyu Jin}
\affiliation{%
  \institution{Dalian Polytechnic University}
  \city{Dalian}
  \country{China}
  }
\email{jiyu.jin@dlpu.edu.cn}






\renewcommand{\shortauthors}{Qiyuan Guan et al.}

\begin{abstract}
Existing all-in-one image restoration approaches, which aim to handle multiple weather degradations within a single framework, are predominantly trained and evaluated using mixed single-weather synthetic datasets.
However, these datasets often differ significantly in resolution, style, and domain characteristics, leading to substantial domain gaps that hinder the development and fair evaluation of unified models.
Furthermore, the lack of a large-scale, real-world all-in-one weather restoration dataset remains a critical bottleneck in advancing this field. 
To address these limitations, we present a real-world all-in-one adverse weather image restoration benchmark dataset, which contains image pairs captured under various weather conditions, including rain, snow, and haze, as well as diverse outdoor scenes and illumination settings.
The resulting dataset provides precisely aligned degraded and clean images, enabling supervised learning and rigorous evaluation.
We conduct comprehensive experiments by benchmarking a variety of task-specific, task-general, and all-in-one restoration methods on our dataset.
Our dataset offers a valuable foundation for advancing robust and practical all-in-one image restoration in real-world scenarios.
The dataset has been publicly released and is available at~\href{https://github.com/guanqiyuan/WeatherBench} {\textit{https://github.com/guanqiyuan/WeatherBench}}.

\end{abstract}




\maketitle

\section{Introduction}

\begin{figure*}[t]
    \centering
    \begin{tabular}{c}
        \includegraphics[width=\linewidth]{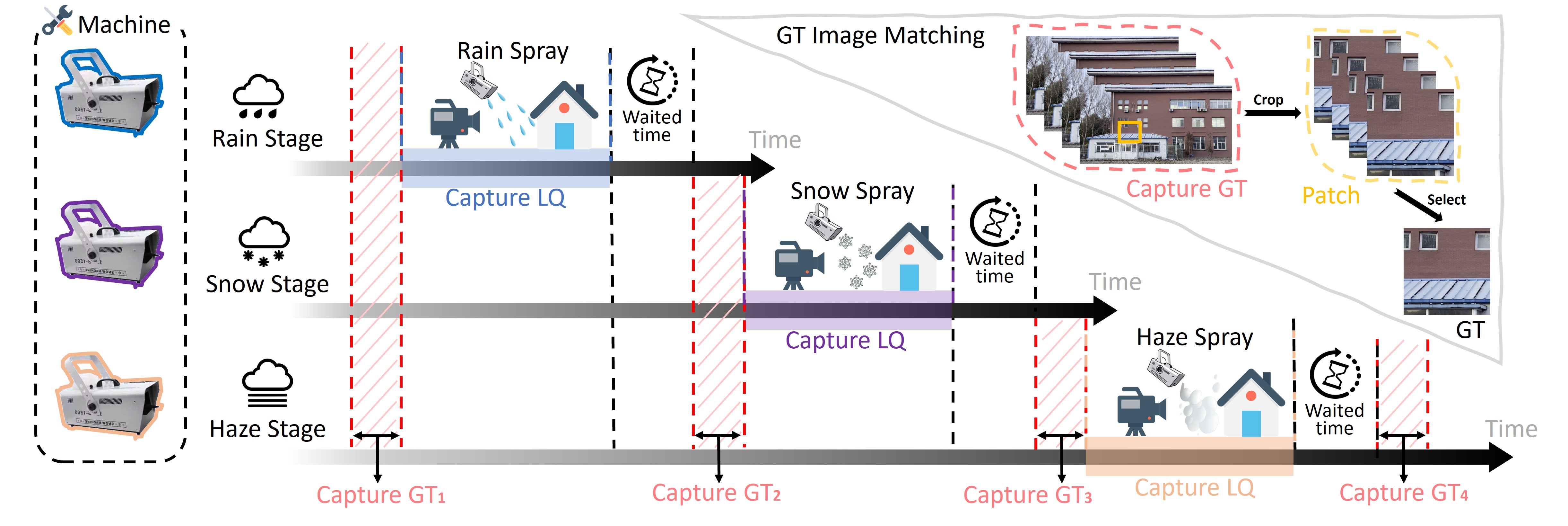}
    \end{tabular}
    \caption{Illustration of the All-in-One Weather Acquisition System (AWAS). By fixing camera settings, we sequentially sprayed and captured images for rain, snow, and haze—scheduling haze last due to its slower dissipation. Each stage involved capturing GT images, capturing LQ images, and waiting for the scene to recover. To ensure high-quality pairs, a GT matching strategy filtered out samples with moving objects.}
    \label{fig:data_pipeine}
\end{figure*}

Visual perception under adverse weather conditions plays a vital role in a wide range of real-world applications \cite{chen2025towards}, including autonomous driving~\cite{caesar2020nuscenes, lou2022testing}, intelligent transportation~\cite{chavhan2022edge}, and outdoor surveillance~\cite{sheng2021surveilling}.
Common weather phenomena such as rain, snow, and haze significantly degrade image quality by introducing occlusions, scattering, and contrast reduction, which in turn undermine the performance of downstream vision tasks such as object detection and tracking~\cite{zhou2022apptracker}.
It is crucial to develop robust all-in-one image restoration techniques capable of effectively removing such degradations to enhance visual clarity and ensure the reliability of high-level perception systems.
However, existing all-in-one approaches~\cite{li2022all, valanarasu2022transweather, sun2024restoring, cui2025adair} are typically trained and evaluated by combining multiple single-weather synthetic datasets~\cite{jiang2020multi, li2018benchmarking, li2019heavy, liu2018desnownet}, which differ significantly in resolution, imaging style, and domain characteristics, making it difficult to ensure consistent and fair evaluation.
Therefore, establishing a unified and realistic benchmark dataset is an urgent necessity for robust multimedia applications under adverse weather conditions.

In recent years, image restoration under adverse weather conditions has been extensively studied as a set of independent sub-tasks, including single image dehazing, deraining, and desnowing.
With the advancement of deep learning, a variety of task-specific methods~\cite{song2023vision, zhang2024depth, chen2023learning, chen2024bidirectional} have been proposed, which are typically trained and evaluated on synthetic datasets~\cite{jiang2020multi, li2018benchmarking, chen2020jstasr} designed to simulate individual weather degradations.
Although promising results have been achieved in controlled environments, these methods often exhibit limited generalization ability in complex real-world scenarios.
To address the limitations of task-specific models, recent research has explored all-in-one image restoration frameworks~\cite{valanarasu2022transweather, potlapalli2023promptir, zhu2024mwformer} that aim to handle multiple degradation types within a single network.
These frameworks have the potential to simplify deployment pipelines and improve the efficiency of practical applications.
However, most existing all-in-one approaches are still developed and evaluated by mixing several single-weather synthetic datasets, which suffer from significant domain gaps.
As a result, the training process is adversely affected, and the resulting models may fail to generalize well to real-world all-in-one weather scenarios.
Moreover, the heavy reliance on artificially synthesized degradations further limits the effectiveness of these methods.
While synthetic data enables controlled experiments and large-scale training, it often fails to capture the complexity and unpredictability of real-world weather conditions—such as the irregular shapes of rain streaks, snow accumulation, or haze scattering under varying illumination.
Incorporating real-world datasets into training can significantly improve performance in practical environments~\cite{li2024foundir}.
However, the limited availability of large-scale, diverse, and realistically degraded weather datasets continues to constrain model performance and hinder further progress in the field.
In particular, the lack of a unified all-in-one weather dataset poses a key bottleneck for developing and fairly evaluating all-in-one models.
Without a consistent benchmark that covers multiple weather types under shared conditions, it is challenging to comprehensively assess model performance or reliably compare different methods.

To this end, we propose an All-in-One Weather Acquisition System (AWAS) specifically designed to capture real-world all-in-one weather-degraded image pairs.
Unlike previous datasets that rely on synthetic degradations or focus on a single type of weather, our dataset is collected under real-world conditions, including rain, snow, and haze, across diverse outdoor scenes and lighting conditions.
Each low-quality (LQ) degraded image is precisely aligned with a clean ground truth (GT) counterpart, enabling supervised learning and quantitative evaluation.
To validate the practical value of our dataset, we conduct comprehensive experiments using task-specific, task-general, and all-in-one methods to evaluate their performance under different weather conditions.

The contributions of our work are summarized as follows:
\begin{itemize}
\item We introduce a real-world all-in-one adverse weather image restoration dataset that includes rain, snow, and haze conditions under diverse scenes and illumination levels.

\item We design an all-in-one weather acquisition system that enables unified collection of images under different real-world weather degradations, avoiding the domain gaps introduced by mixing different single-weather synthetic datasets.

\item We conduct a comprehensive study and evaluation of task-specific methods, task-general methods, and all-in-one methods using our proposed benchmark dataset.
\end{itemize}

\begin{figure*}[t]
    \centering
    \setlength{\tabcolsep}{1mm}
    \begin{tabular}{ccc}
        \includegraphics[width=0.3\linewidth]{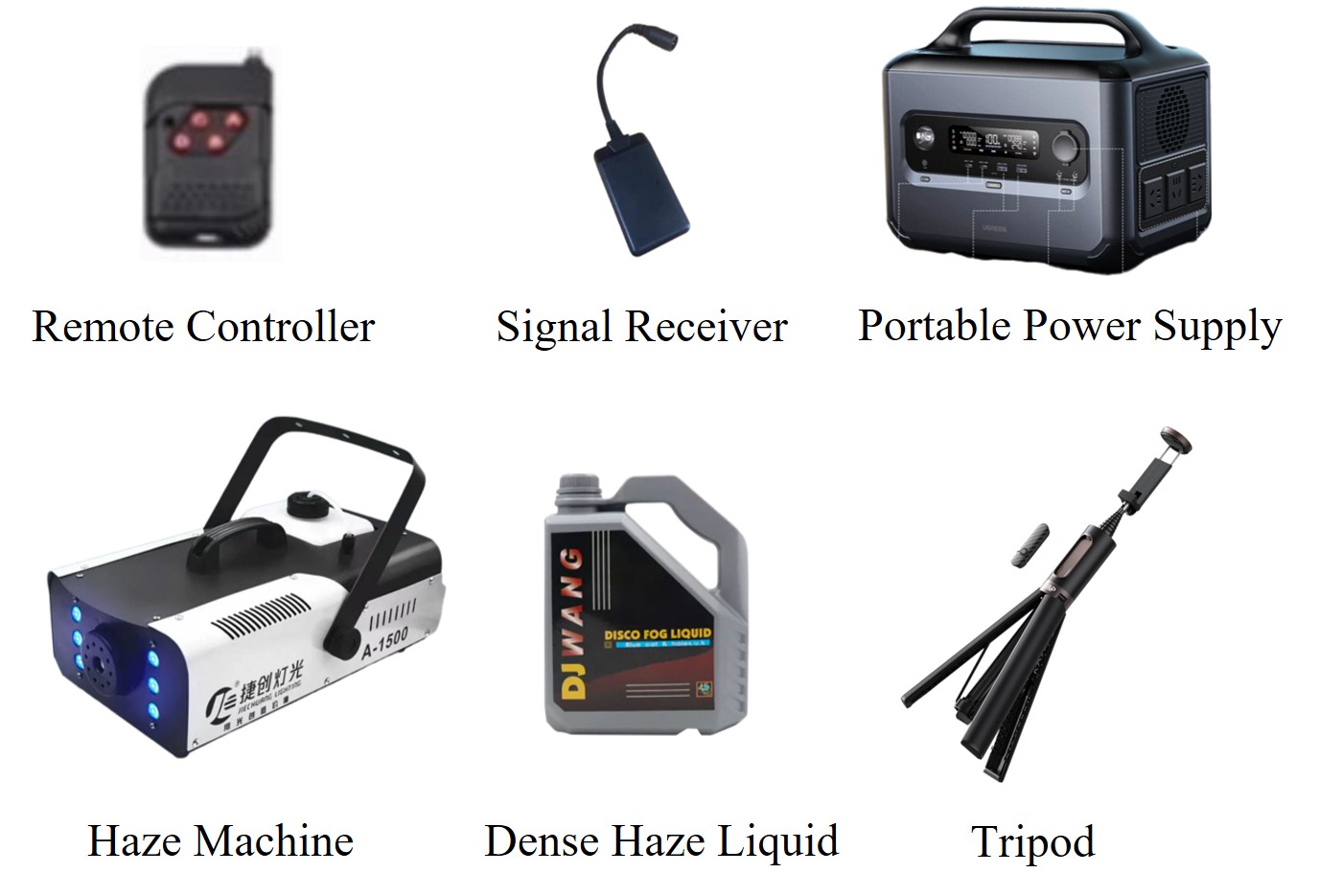} & 
        \includegraphics[width=0.25\linewidth]{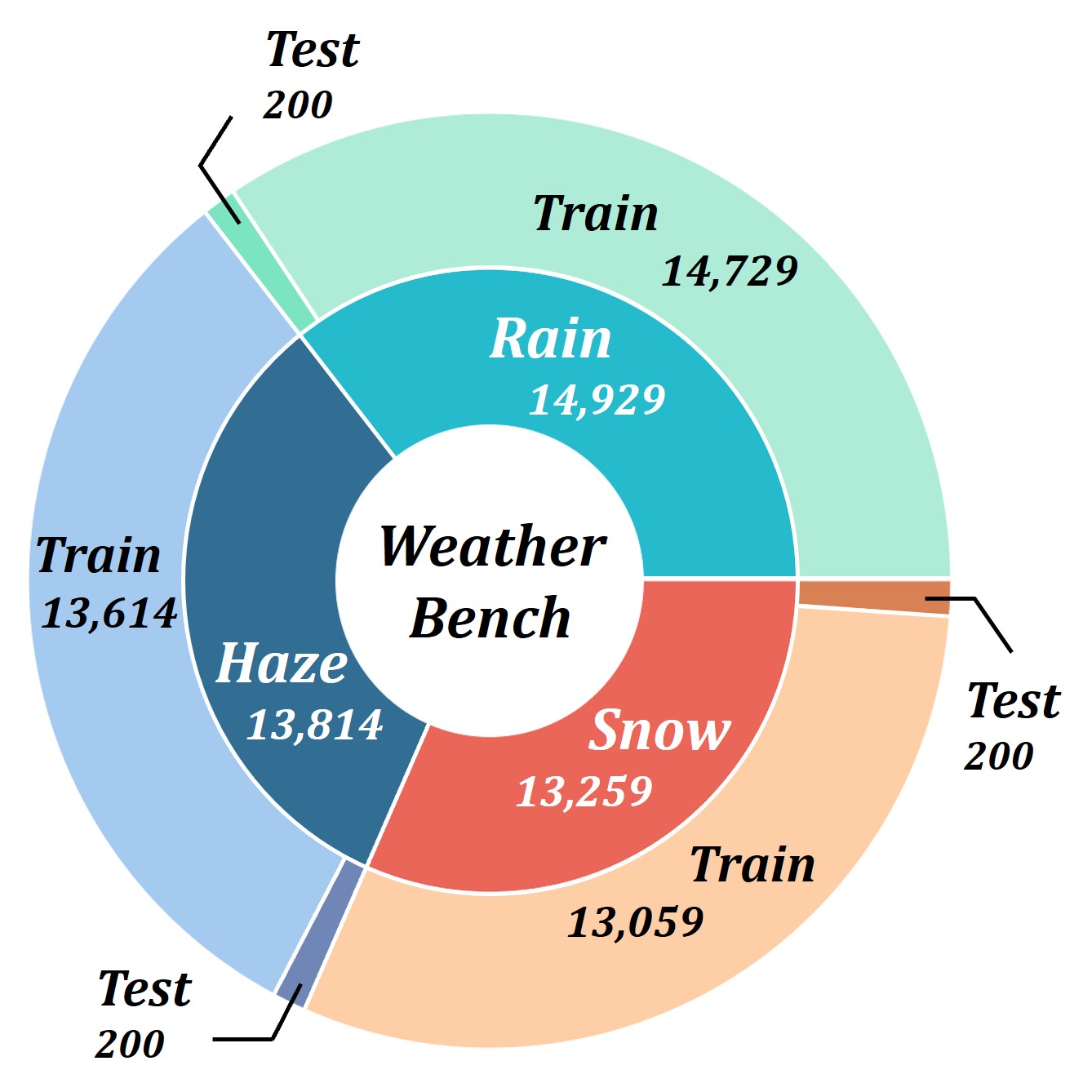} & 
        \includegraphics[width=0.41\linewidth]{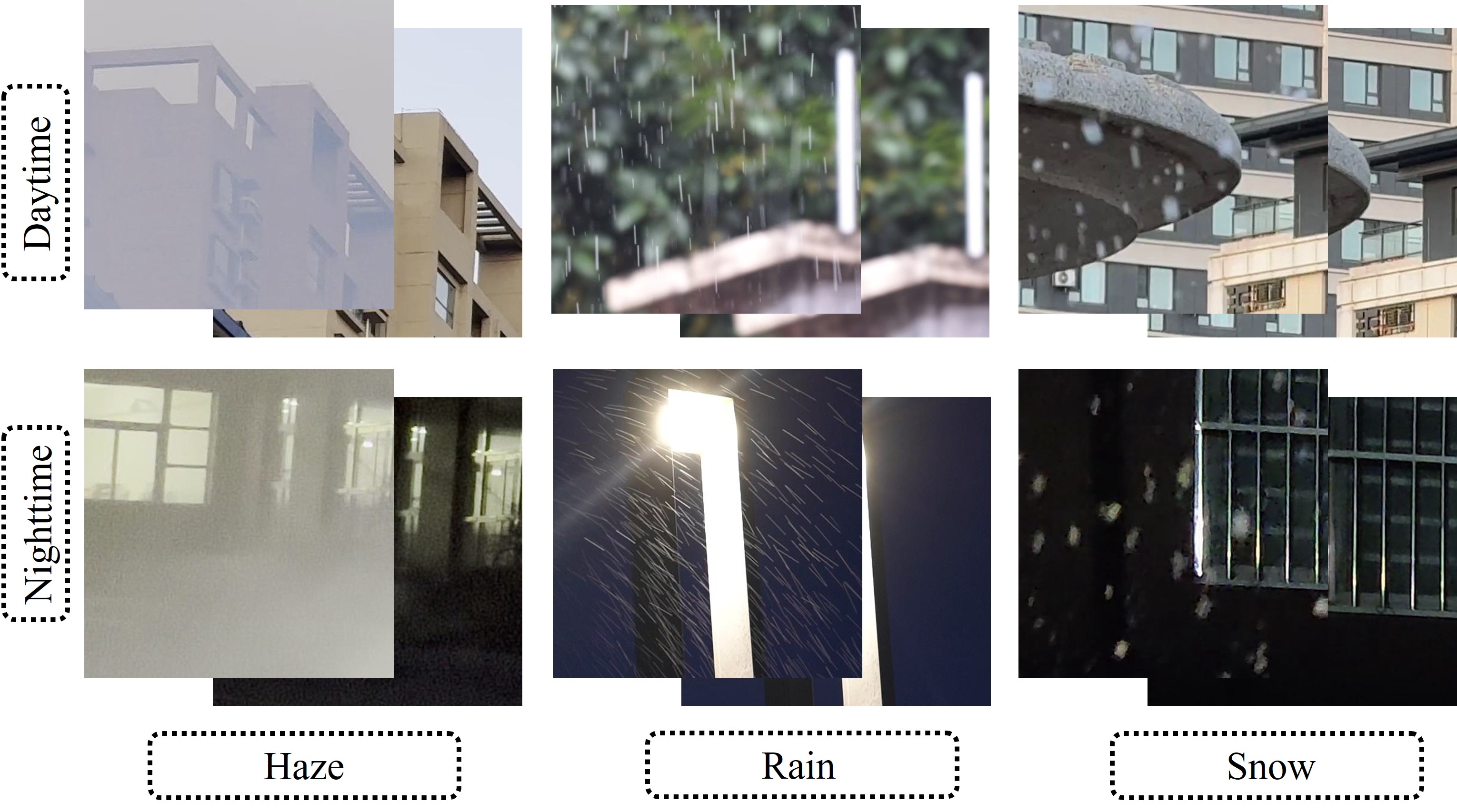} \\
        (a) Equipment  & (b) Distribution & (c) Samples
    \end{tabular}
    \caption{Illustration of the proposed WeatherBench dataset.~(a) Overview of the data collection equipment (taking haze as an example).~(b) Data distribution of the proposed dataset.~(c) Sample image pairs from our dataset. In each pair, the left image shows the LQ, while the right image displays the corresponding GT image.~We collect images with different degradation types under both daytime and nighttime scenes.}
    \label{fig:data_show}
\end{figure*}

\section{Related Work}
\subsection{Single-Weather Image Restoration}
Image restoration under adverse weather has long been divided into separate tasks such as dehazing, deraining, and desnowing. Most deep learning methods are trained and evaluated on synthetic single-weather datasets.
In the \textbf{image deraining} field, early methods such as JORDER~\cite{yang2019joint} focused on learning spatial priors to remove rain streaks, while later approaches like PReNet~\cite{ren2019progressive}, NeRD-Rain~\cite{chen2024bidirectional}, and MSDT~\cite{chen2024rethinking} employed recurrent and multi-scale architectures to improve performance.
Common datasets include Rain13k~\cite{jiang2020multi} and Outdoor-Rain~\cite{li2019heavy}, where rain streaks are synthetically added to clean images.
To explore model performance on real-world rain removal, datasets such as SPA-Data~\cite{wang2019spatial}, RealRain1k~\cite{li2022toward}, MPID~\cite{li2019single}, and RainDS~\cite{quan2021removing} include partially real rainy scenes.
In the \textbf{image desnowing} domain, Snow100K~\cite{liu2018desnownet}, SRRS~\cite{chen2020jstasr}, and CSD~\cite{chen2021all} have become benchmark datasets, giving rise to models such as DesnowNet~\cite{liu2018desnownet}, JSTASR~\cite{chen2020jstasr}, and HDCWNet~\cite{chen2021all} to tackle snow particles of varying densities and scales.
Meanwhile, \textbf{image dehazing} methods like AOD-Net~\cite{li2017aod}, FDU~\cite{dong2020physics}, and DehazeFormer~\cite{song2023vision} integrate physical priors and learning strategies and have achieved notable progress on datasets such as RESIDE~\cite{li2018benchmarking} and DenseHaze~\cite{ancuti2019dense}.
Despite these advances, the synthetic nature of many datasets limits the generalization ability of models in real-world environments, where weather degradations are often more complex and unpredictable.

\subsection{All-in-One Weather Image Restoration}
To address real-world challenges, recent studies explore all-in-one models that handle multiple weather degradations in a unified framework.
Some task-general methods, such as MPRNet~\cite{zamir2021multi}, Restormer~\cite{zamir2022restormer}, and MambaIR~\cite{guo2024mambair}, although not specifically designed for adverse weather degradations, have demonstrated strong generalization capabilities across various weather-related datasets.
%
%
All-in-one~\cite{li2020all} adopts a generator architecture composed of multiple task-specific encoders and a shared decoder, enabling unified restoration under various weather conditions.
Transformer-based methods like TransWeather~\cite{valanarasu2022transweather} employ independent encoder-decoder structures to handle degradations such as rain, snow, and haze.
WGWS-Net~\cite{zhu2023learning} learns both weather-general and weather-specific representations in two separate stages.
Recently, diffusion-based methods such as WeatherDiffusion~\cite{ozdenizci2023restoring} and DiffUIR~\cite{zheng2024selective} have further broadened the research directions in this area.

However, existing all-in-one weather image restoration research still faces a major bottleneck: the lack of a sufficiently large and realistically degraded benchmark dataset.
While datasets like Multi-Weather6k~\cite{li2024rethinking} cover rain, snow, and haze, they are limited in scale and rely entirely on synthetic degradation, making them inadequate for model training.
Due to the absence of a unified all-in-one weather benchmark, most all-in-one models are still trained and evaluated by combining multiple single-weather datasets.
However, significant differences in resolution, capture devices, and image styles among these datasets often lead to domain gaps, hindering the development and fair evaluation of all-in-one models.
To address this issue, we introduce a real-world all-in-one weather image restoration dataset that includes diverse weather conditions, lighting scenarios, and degradation intensities.
This dataset aims to provide a unified platform for training and evaluating all-in-one weather restoration models, advancing the field toward more general and practical solutions.

\begin{table}[!t]
\footnotesize
\setlength{\tabcolsep}{4pt}
\caption{Comparison of existing weather degradation datasets.~Our dataset covers a wider range of degradation types and provides a significantly larger number of image pairs.~"Real" denotes real-world data.}
\begin{tabular}{c|c|ccc|cc|c}
\hline
\multirow{2}{*}{Dataset} & \multirow{2}{*}{\makecell{Image \\ pairs}} & \multicolumn{3}{c|}{Weather types}  & \multicolumn{2}{c|}{Time} & \multirow{2}{*}{Real} \\ \cline{3-5} \cline{6-7}
                         &                                             & Haze            & Rain            & Snow                     & Day             & Night          &         \\ \hline
DenseHaze \cite{ancuti2019dense}                & 55                                          & \CheckmarkBold  &                 &                      & \CheckmarkBold  &                      & \CheckmarkBold  \\
RESIDE \cite{li2018benchmarking}                   & 13,990                                      & \CheckmarkBold  &                 &                      & \CheckmarkBold  &                   & \CheckmarkBold  \\
Outdoor-Rain \cite{li2019heavy}             & 10,500                                      &                 & \CheckmarkBold  &                       & \CheckmarkBold  & \CheckmarkBold     &   \\
Rain13k \cite{jiang2020multi}                  & 18,012                                      &                 & \CheckmarkBold  &                       & \CheckmarkBold  & \CheckmarkBold    &  \\
SRRS \cite{chen2020jstasr}                    & 15,005                                      &                 &                 & \CheckmarkBold       & \CheckmarkBold  &                &    \\
CSD \cite{chen2021all}                      & 10,000                                      &                 &                 & \CheckmarkBold       & \CheckmarkBold  &           &   \\
GT-RAIN \cite{ba2022not}                  & 30,317                                      &                 & \CheckmarkBold  &                     & \CheckmarkBold  &            & \CheckmarkBold    \\
Multi-Weather6k \cite{li2024rethinking}          & 6,000                                       & \CheckmarkBold  & \CheckmarkBold  & \CheckmarkBold        & \CheckmarkBold  &        &          \\
\textbf{WeatherBench (Ours)}          & 42,002                                      & \CheckmarkBold  & \CheckmarkBold  & \CheckmarkBold       & \CheckmarkBold  & \CheckmarkBold    & \CheckmarkBold      \\ \hline
\end{tabular}
\label{tab:dataset_survey}
\end{table}

\begin{table*}[t]
\footnotesize
\renewcommand{\arraystretch}{1.3}
\setlength{\tabcolsep}{3pt}
\caption{Quantitative comparison of different categories of methods on the WeatherBench dataset.~The best and second-best values are \textbf{blod} and \underline{underlined}.}
\begin{tabular}{cc|c|c|ccc|ccc|ccc|ccc}
\hline
\multicolumn{2}{c|}{\multirow{2.5}{*}{Type}}                                            & \multirow{2}{*}{Method} & \multirow{2}{*}{Venue} & \multicolumn{3}{c|}{Dehaze} & \multicolumn{3}{c|}{Derain} & \multicolumn{3}{c|}{Desnow} & \multicolumn{3}{c}{Average} \\ \cline{5-16} 
\multicolumn{2}{c|}{}                                                                 &                         &                        & PSNR$\uparrow$     & SSIM$\uparrow$    & LPIPS$\downarrow$  & PSNR$\uparrow$     & SSIM$\uparrow$    & LPIPS$\downarrow$  & PSNR$\uparrow$     & SSIM$\uparrow$    & LPIPS$\downarrow$  & PSNR$\uparrow$     & SSIM$\uparrow$    & LPIPS$\downarrow$  \\ \hline
\multicolumn{1}{c|}{\multirow{5}{*}{\makecell{Task-Specific \\ Methods}}} & \multirow{2}{*}{Dehaze} & DehazeFormer~\cite{song2023vision}            & TIP 2023               & \underline{24.1151}  & \underline{0.7447}  & 0.3454 & 36.0537  & 0.9539  & 0.1810 & 28.8771  & 0.8492  & 0.1782 & \underline{29.6820}  & 0.8493  & 0.2349 \\
\multicolumn{1}{c|}{}                                       &                         & DCMPNet~\cite{zhang2024depth}              & CVPR 2024               & 21.1808   & 0.5056  & 0.4914 & 32.0352   & 0.8756  & 0.2817 & 24.8052   & 0.6135  & 0.5455 & 26.0071   & 0.6649  & 0.4395 \\ \cline{2-16} 
\multicolumn{1}{c|}{}                                       & \multirow{2}{*}{Derain} & DRSformer~\cite{chen2023learning}               & CVPR 2023              & 19.9543  & 0.6944  & 0.4035 & 33.9784  & 0.9426  & 0.2092 & 27.9988  & 0.8358  & 0.1966 & 27.3105  & 0.8243  & 0.2698 \\
\multicolumn{1}{c|}{}                                       &                         & NeRD-Rain~\cite{chen2024bidirectional}               & CVPR 2024              & 21.5155  & 0.7184  & 0.3859 & 35.7362  & 0.9503  & 0.1815 & 28.8665  & 0.8507  & 0.1823 & 28.7061  & 0.8398  & 0.2499 \\ \cline{2-16} 
\multicolumn{1}{c|}{}                                       & Desnow                  & SnowFormer~\cite{chen2022snowformer}            & arXiv 2022                      & 22.7099   & 0.7362  & \underline{0.3049} & 35.1831   & 0.9514  & 0.1551 & 29.3002   & 0.8678  & \underline{0.1425} & 29.0644   & 0.8518  & \underline{0.2008} \\ \hline
\multicolumn{2}{c|}{\multirow{2}{*}{\makecell{Task-General \\ Methods}}}                            & MPRNet~\cite{zamir2021multi}                  & CVPR 2021              & 23.2662  & 0.7393  & 0.3551 & \underline{36.1419}  & 0.9537  & 0.1709 & 29.1765  & 0.8598  & 0.1766 & 29.5282  & 0.8509  & 0.2342 \\
\multicolumn{2}{c|}{}                                                                 & Restormer~\cite{zamir2022restormer}               & CVPR 2022              & 19.3049  & 0.6866  & 0.4120 & 34.4895  & 0.9448  & 0.1970 & 27.9500  & 0.8360  & 0.1971 & 27.2481  & 0.8225  & 0.2687 \\ \hline
\multicolumn{2}{c|}{\multirow{8}{*}{\makecell{All-in-One \\ Methods}}}                              & AirNet~\cite{li2022all}                  & CVPR 2022              & 20.9399  & 0.7054  & 0.3829 & 33.5855  & 0.9418  & 0.2237 & 22.0594  & 0.7799  & 0.2912 & 25.5283  & 0.8090  & 0.2993 \\
\multicolumn{2}{c|}{}                                                          & TransWeather~\cite{valanarasu2022transweather}            & CVPR 2022              & 19.7904  & 0.6800  & 0.3969 & 29.3372  & 0.9034  & 0.2943 & 24.9631  & 0.7959  & 0.2309 & 24.6969  & 0.7931  & 0.3074 \\
\multicolumn{2}{c|}{}                                                                 & PromptIR~\cite{potlapalli2023promptir}                & NeurIPS 2023           & 21.1132  & 0.7128  & 0.3751 & 34.5374  & 0.9443  & 0.1980 & 27.9270  & 0.8363  & 0.1951 & 27.8592  & 0.8311  & 0.2561 \\
\multicolumn{2}{c|}{}                                                                 & WGWS-Net~\cite{zhu2023learning}                & CVPR 2023              & 13.7943  & 0.6027  & 0.5351 & \textbf{37.0770}  & \textbf{0.9608}  & \textbf{0.1169} & 20.8051  & 0.7799  & 0.2481 & 23.8921  & 0.7811  & 0.3000 \\
\multicolumn{2}{c|}{}                                                                 & DiffUIR~\cite{zheng2024selective}                 & CVPR 2024              & 22.7374  & 0.7439  & 0.3550 & 35.9333  & \underline{0.9547}  & 0.1720 & \underline{29.5040}  & \underline{0.8696}  & 0.1619 & 29.3916  & \underline{0.8561}  & 0.2296 \\
\multicolumn{2}{c|}{}                                                                 & MWFormer~\cite{zhu2024mwformer}                & TIP 2024               & \textbf{24.4223}  & \textbf{0.7462}  & \textbf{0.2837} & 35.1464  & 0.9507  & \underline{0.1532} & \textbf{29.9783}  & \textbf{0.8721}  & \textbf{0.1325} & \textbf{29.8490}  & \textbf{0.8563}  & \textbf{0.1898} \\
\multicolumn{2}{c|}{}                                                                 & Histoformer~\cite{sun2024restoring}             & ECCV 2024              & 17.6861  & 0.6691  & 0.4371 & 30.6981  & 0.9160  & 0.2786 & 25.3939  & 0.8076  & 0.2252 & 24.5927  & 0.7976  & 0.3136 \\
\multicolumn{2}{c|}{}                                                                 & AdaIR~\cite{cui2025adair}                   & ICLR 2025              & 23.0837  & 0.7311  & 0.3505 & 34.8685  & 0.9458  & 0.1917 & 28.4436  & 0.8372  & 0.1790 & 28.7986  & 0.8380  & 0.2404 \\ \hline
\end{tabular}
\label{tab:methods}
\end{table*}

\section{WeatherBench Dataset}

\subsection{All-in-One Weather Acquisition System}
To collect a large-scale dataset of real-world weather-degraded image pairs, we develop an All-in-One Weather Acquisition System (AWAS), as illustrated in Figure~\ref{fig:data_pipeine}.
This system utilizes a haze generator (JIECHUANG LIGHTING A-1500), a rain generator (Leike 1500), and a snow generator (S-1500) to simulate various weather degradations.
A 1200W portable power station supplies electricity to all components, and images are captured using a SONY ILCE-7M3 camera mounted on a tripod for stability.
To minimize interference during the data acquisition process, we employ a wireless remote controller and signal receiver for remote operation.
The experimental equipment is illustrated in Figure~\ref{fig:data_show} (a).

\subsection{Data Collection}
{\flushleft\textbf{Implementation details}.}~
%
To ensure data realism, we collect images under overcast conditions to maintain consistent brightness and texture across pairs. To reduce misalignment caused by scene changes or object motion, we carefully select static environments with minimal wind, low pedestrian flow, and no dynamic objects, such as architectural scenes.
%
After capturing each weather condition, we wait for the scene to return to its original state before acquiring clear images. Since haze takes longer to dissipate and may cause scene changes, we schedule its collection after rain and snow. To maintain image quality, we clean the camera lens regularly. These steps ensure accurate alignment for reliable model training and evaluation.

\begin{figure*}[t]
    \centering
    \setlength{\tabcolsep}{0.5pt}
    \renewcommand{\arraystretch}{0.7}
    \begin{tabular}{cccccccccc} 
        \includegraphics[width=0.1\textwidth]{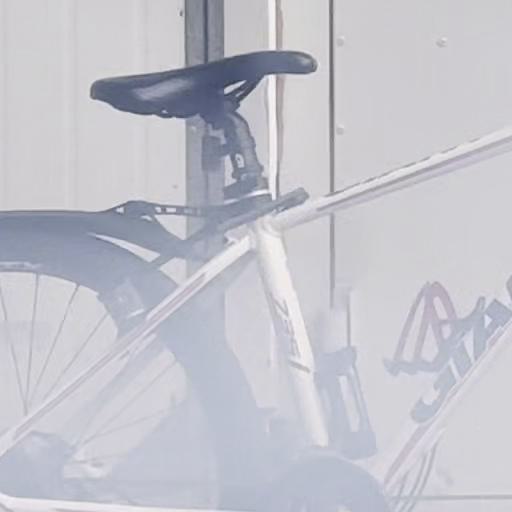} &
        \includegraphics[width=0.1\textwidth]{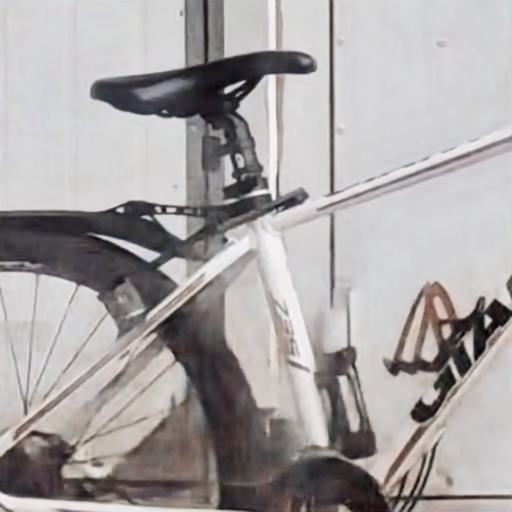} & 
        \includegraphics[width=0.1\textwidth]{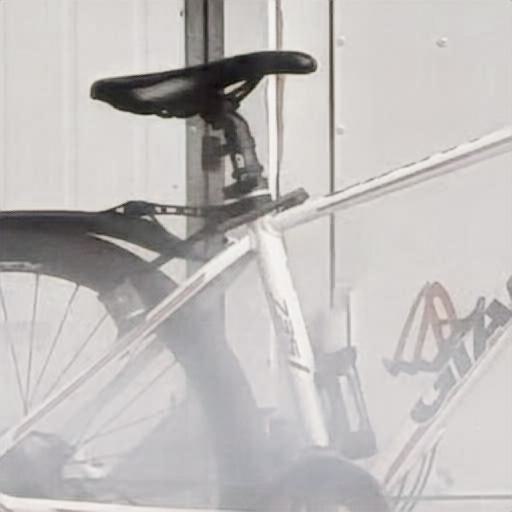} & 
        \includegraphics[width=0.1\textwidth]{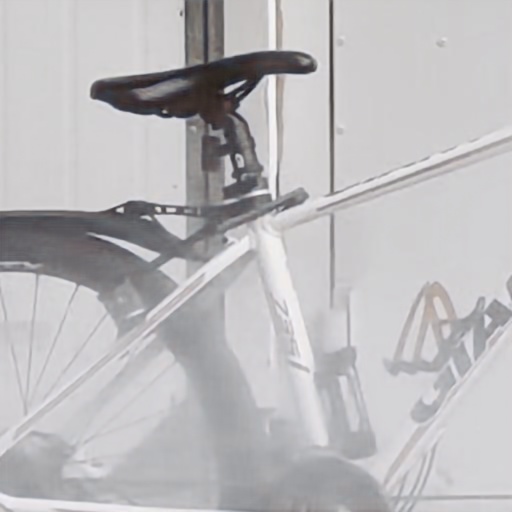} & 
        \includegraphics[width=0.1\textwidth]{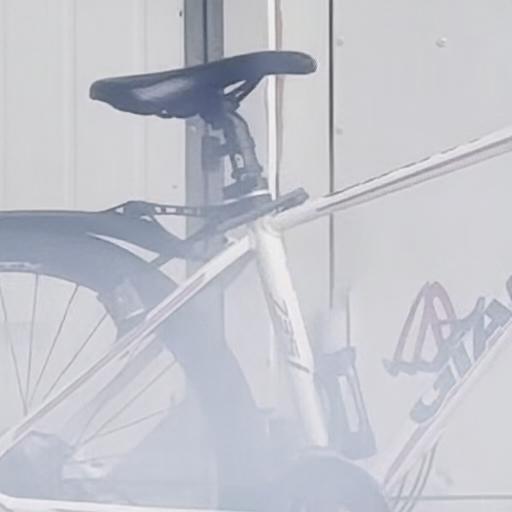} &
        \includegraphics[width=0.1\textwidth]{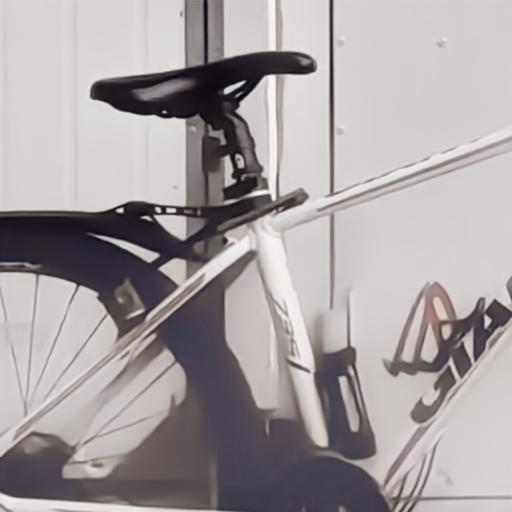} & 
        \includegraphics[width=0.1\textwidth]{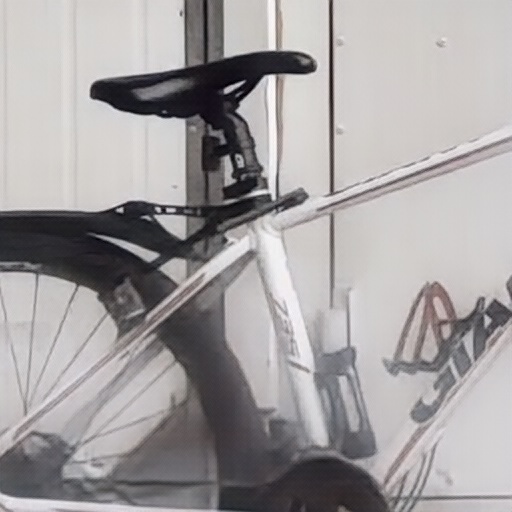} &
        \includegraphics[width=0.1\textwidth]{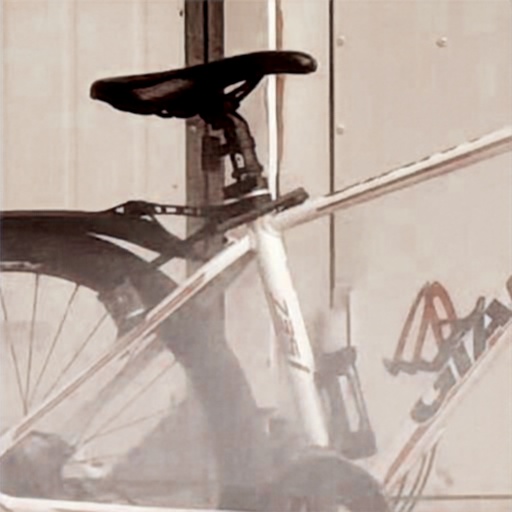} &
        \includegraphics[width=0.1\textwidth]{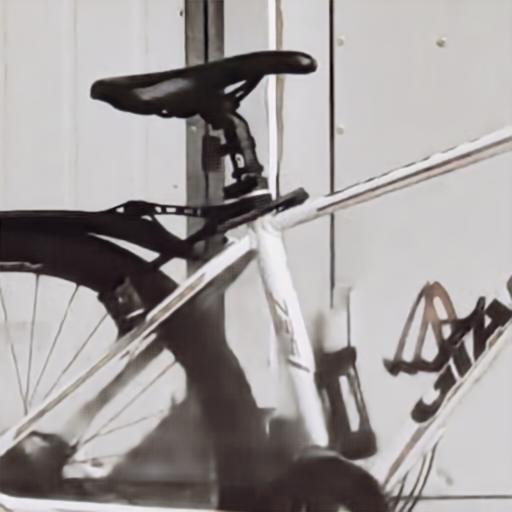} &
        \includegraphics[width=0.1\textwidth]{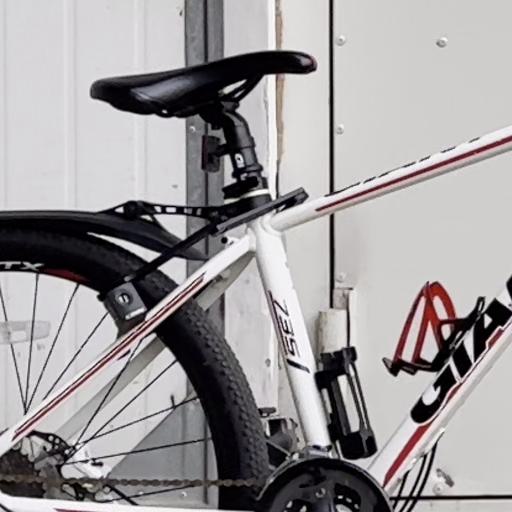} \\

        \includegraphics[width=0.1\textwidth]{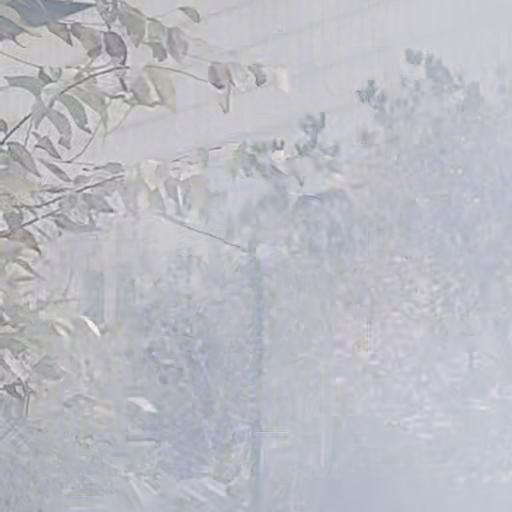} &
        \includegraphics[width=0.1\textwidth]{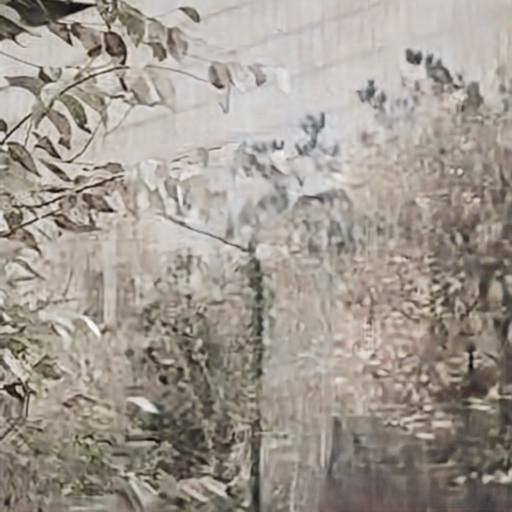} & 
        \includegraphics[width=0.1\textwidth]{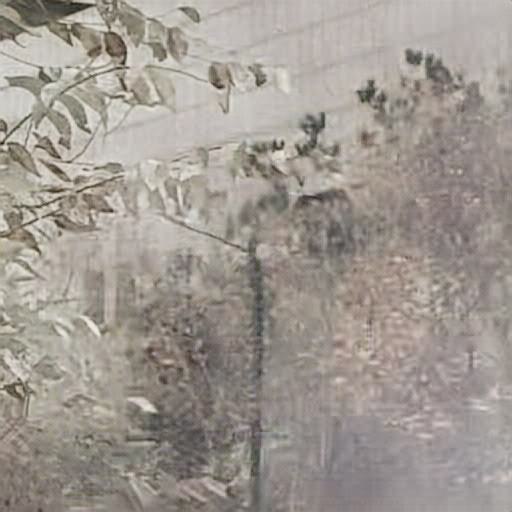} & 
        \includegraphics[width=0.1\textwidth]{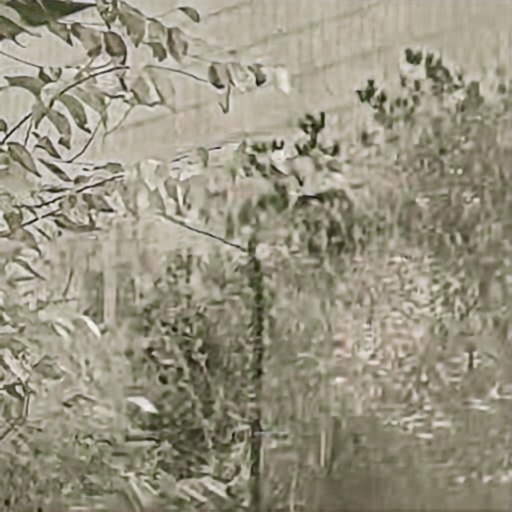} & 
        \includegraphics[width=0.1\textwidth]{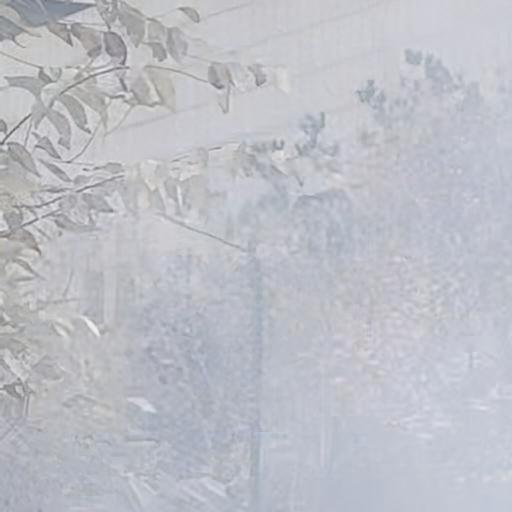} &
        \includegraphics[width=0.1\textwidth]{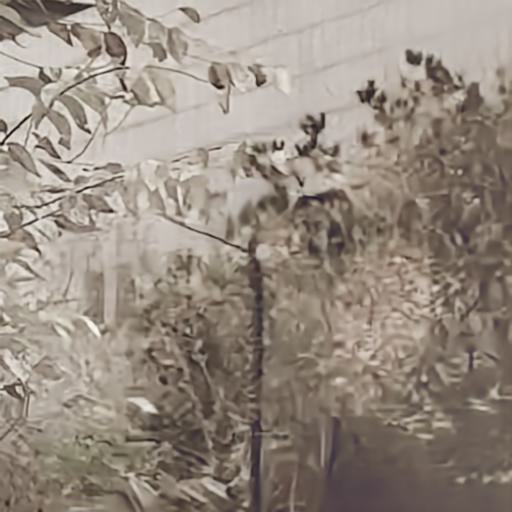} & 
        \includegraphics[width=0.1\textwidth]{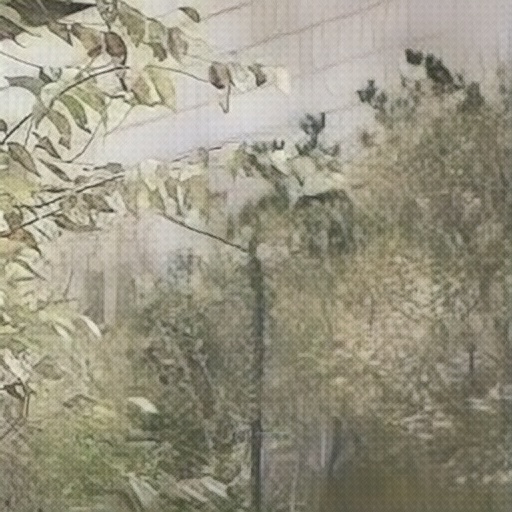} &
        \includegraphics[width=0.1\textwidth]{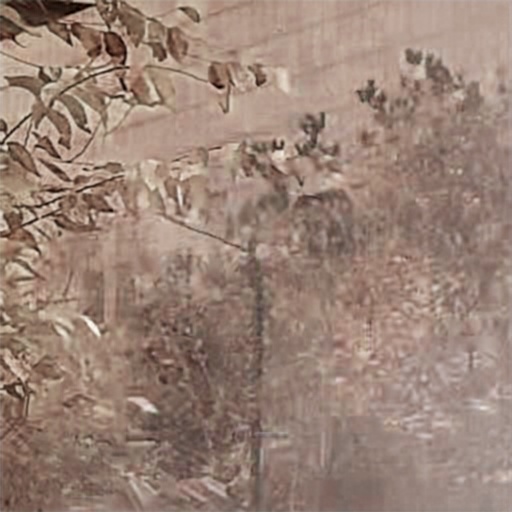} &
        \includegraphics[width=0.1\textwidth]{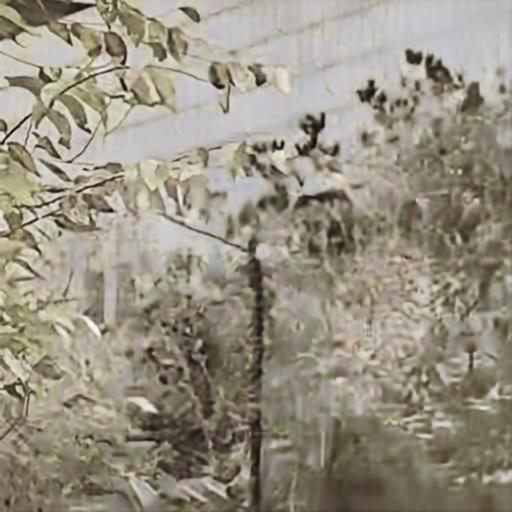} &
        \includegraphics[width=0.1\textwidth]{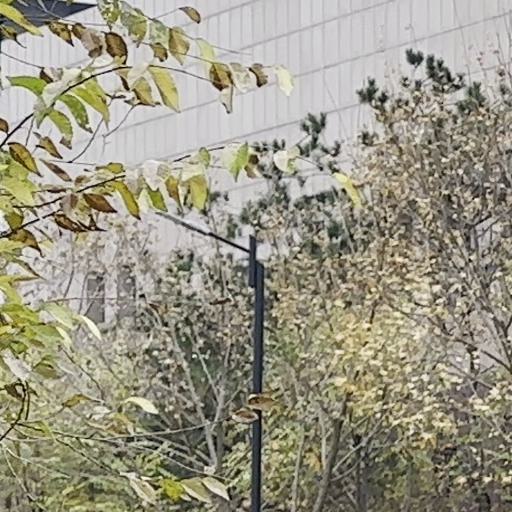} \\

        \includegraphics[width=0.1\textwidth]{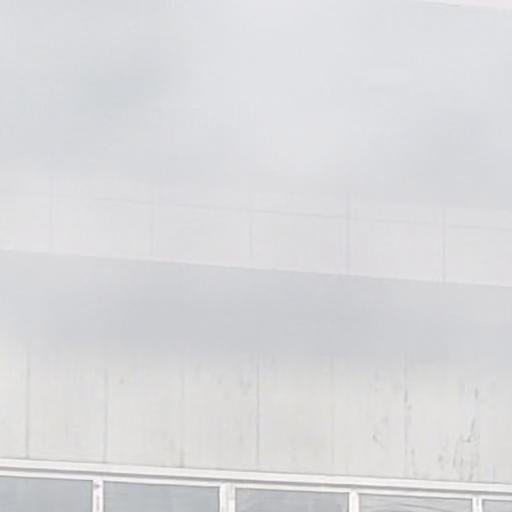} &
        \includegraphics[width=0.1\textwidth]{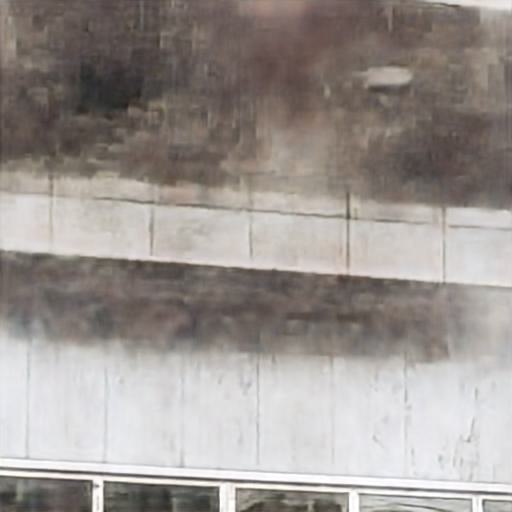} & 
        \includegraphics[width=0.1\textwidth]{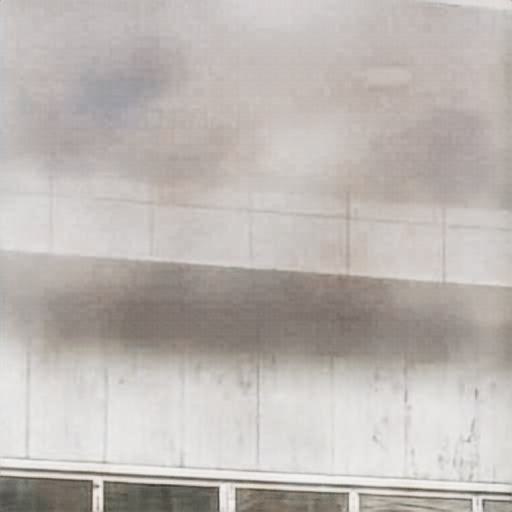} & 
        \includegraphics[width=0.1\textwidth]{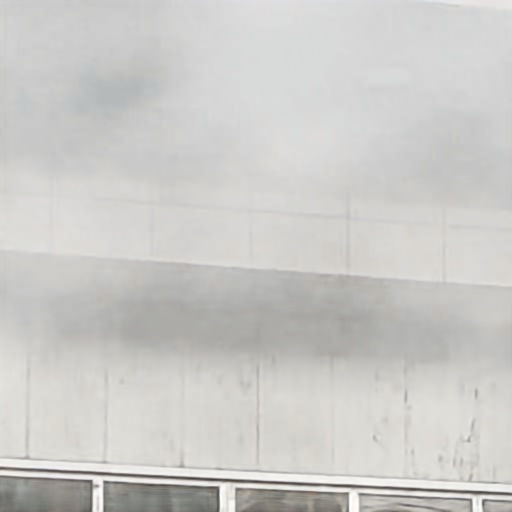} & 
        \includegraphics[width=0.1\textwidth]{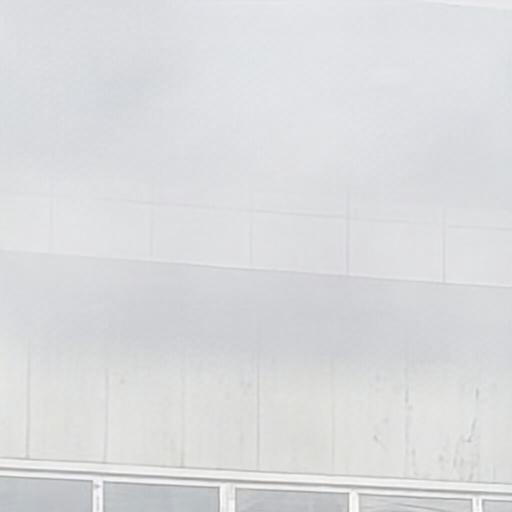} &
        \includegraphics[width=0.1\textwidth]{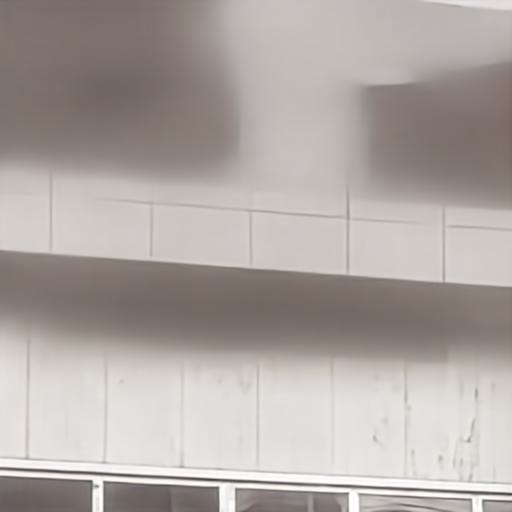} & 
        \includegraphics[width=0.1\textwidth]{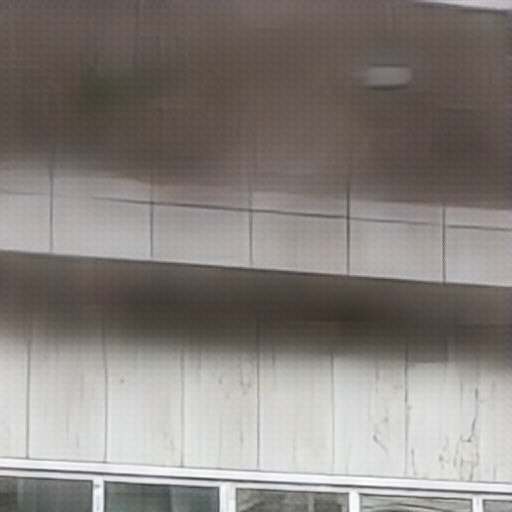} &
        \includegraphics[width=0.1\textwidth]{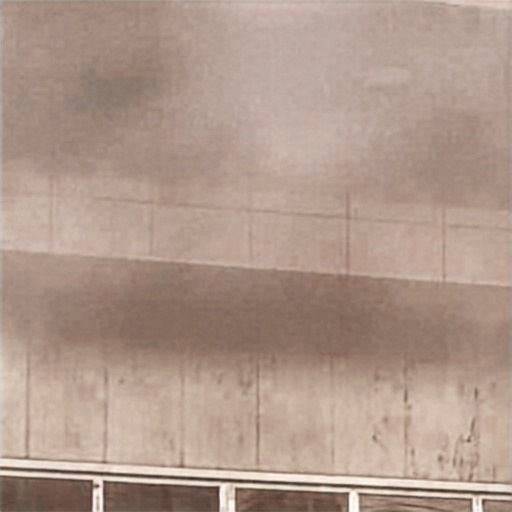} &
        \includegraphics[width=0.1\textwidth]{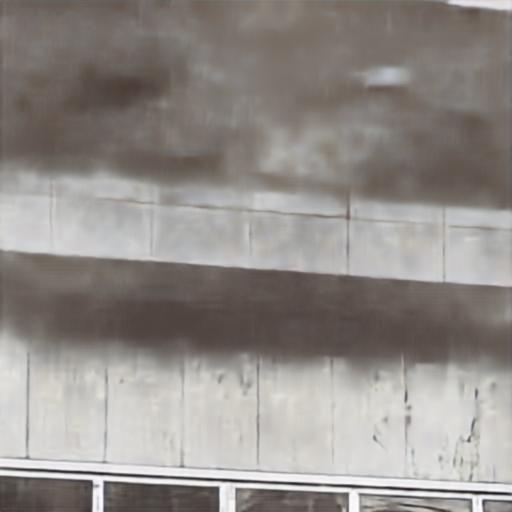} &
        \includegraphics[width=0.1\textwidth]{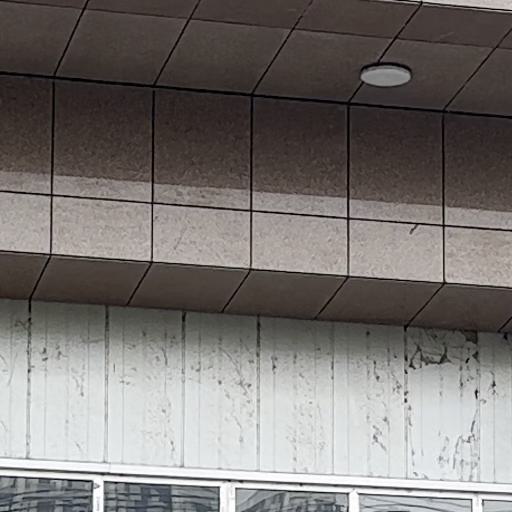} \\
        
        \includegraphics[width=0.1\textwidth]{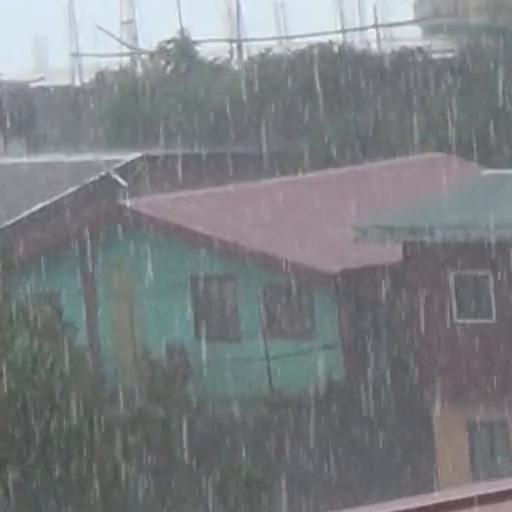} &
        \includegraphics[width=0.1\textwidth]{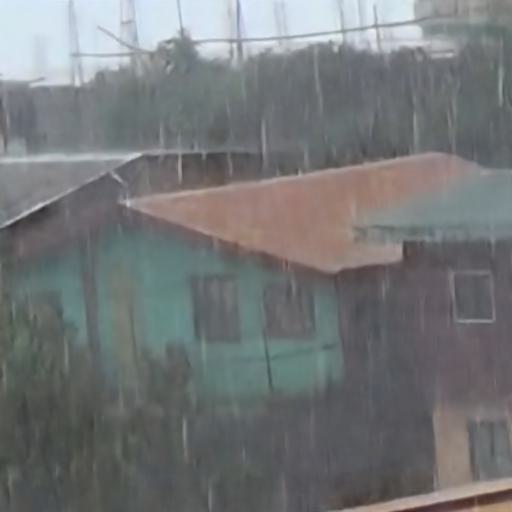} & 
        \includegraphics[width=0.1\textwidth]{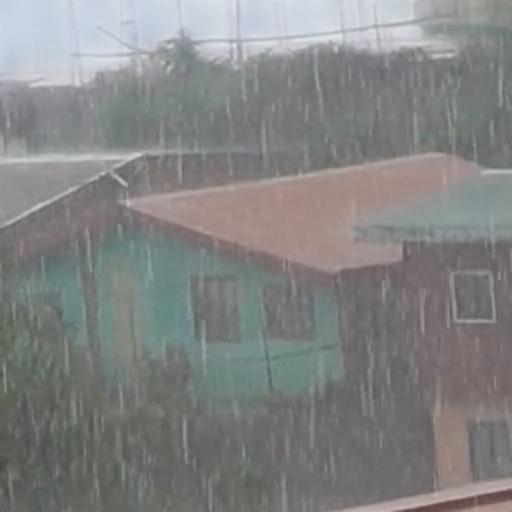} & 
        \includegraphics[width=0.1\textwidth]{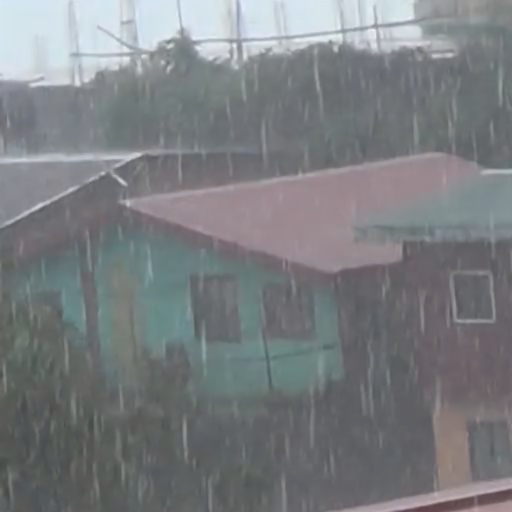} & 
        \includegraphics[width=0.1\textwidth]{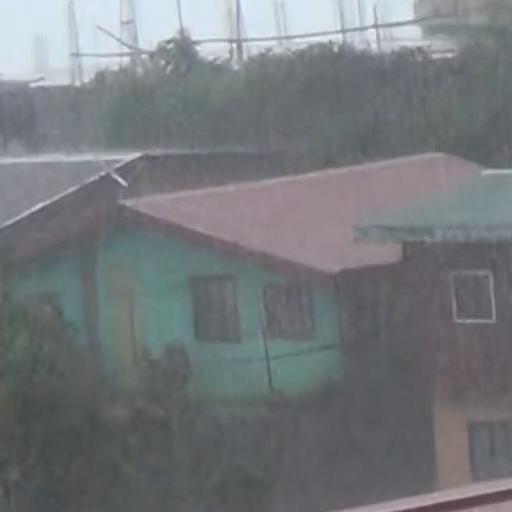} &
        \includegraphics[width=0.1\textwidth]{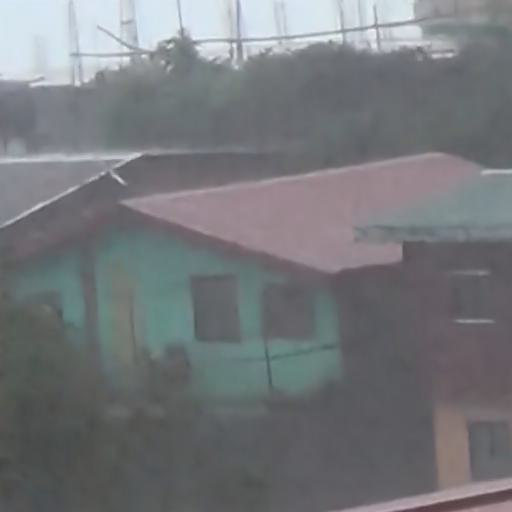} & 
        \includegraphics[width=0.1\textwidth]{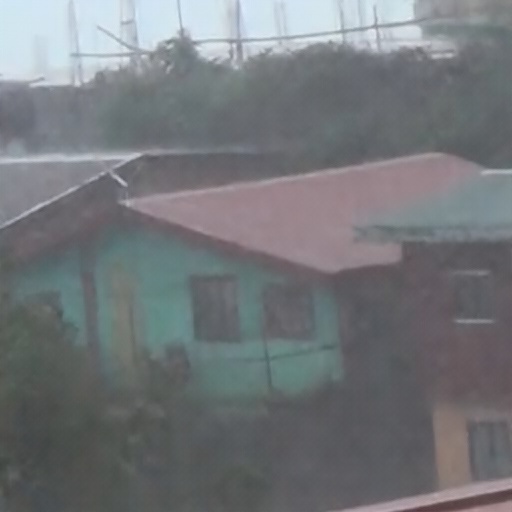} &
        \includegraphics[width=0.1\textwidth]{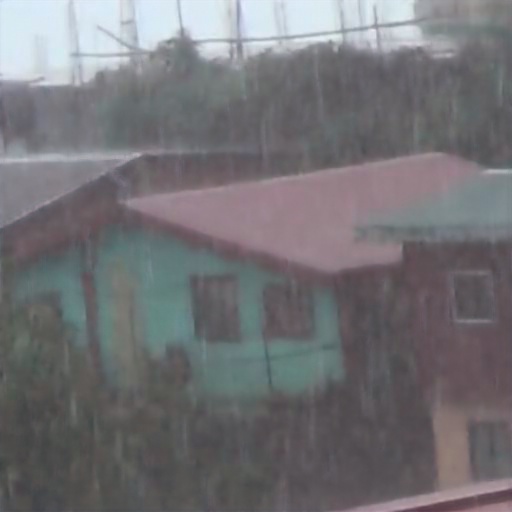} &
        \includegraphics[width=0.1\textwidth]{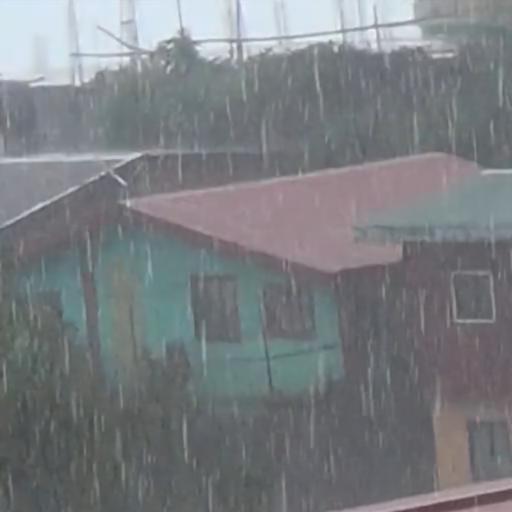} &
        \includegraphics[width=0.1\textwidth]{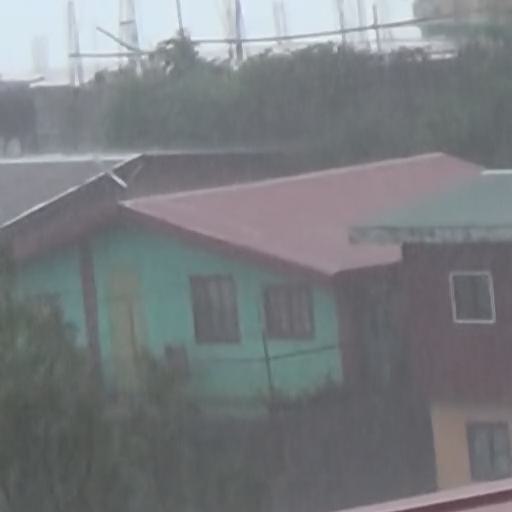} \\

        \includegraphics[width=0.1\textwidth]{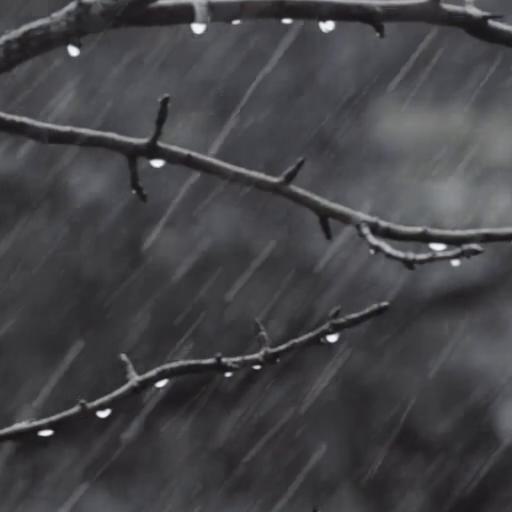} &
        \includegraphics[width=0.1\textwidth]{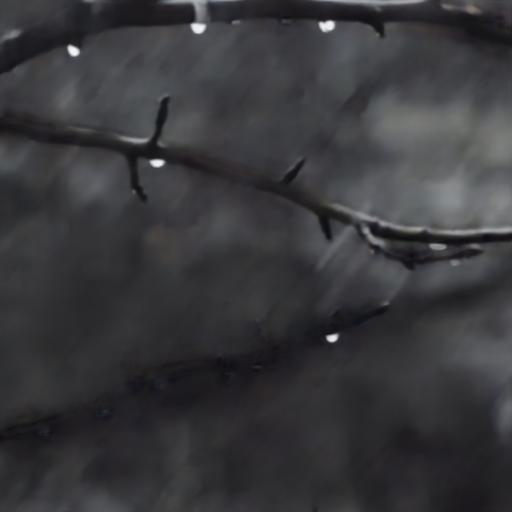} & 
        \includegraphics[width=0.1\textwidth]{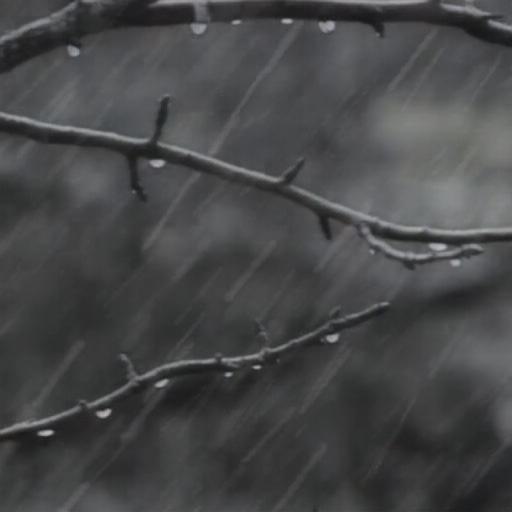} & 
        \includegraphics[width=0.1\textwidth]{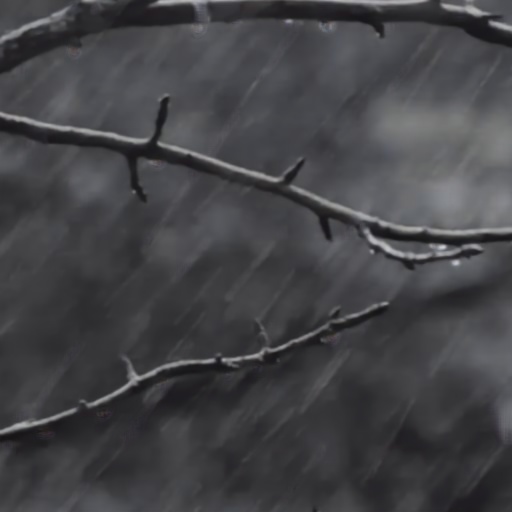} & 
        \includegraphics[width=0.1\textwidth]{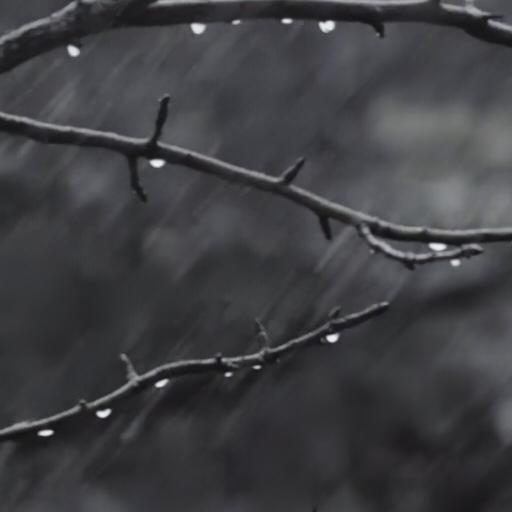} &
        \includegraphics[width=0.1\textwidth]{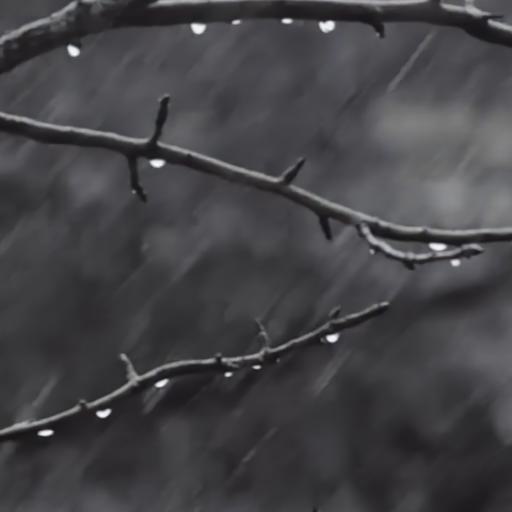} & 
        \includegraphics[width=0.1\textwidth]{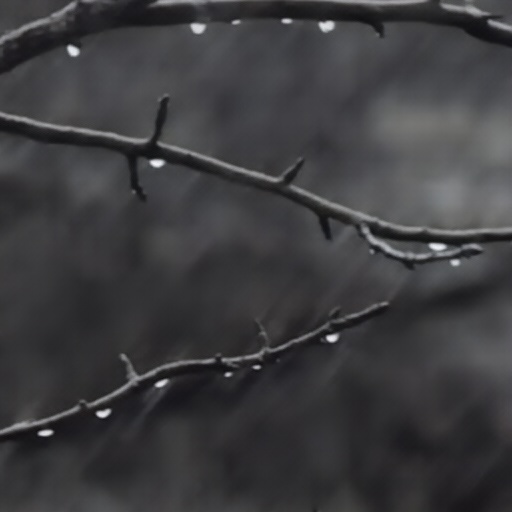} &
        \includegraphics[width=0.1\textwidth]{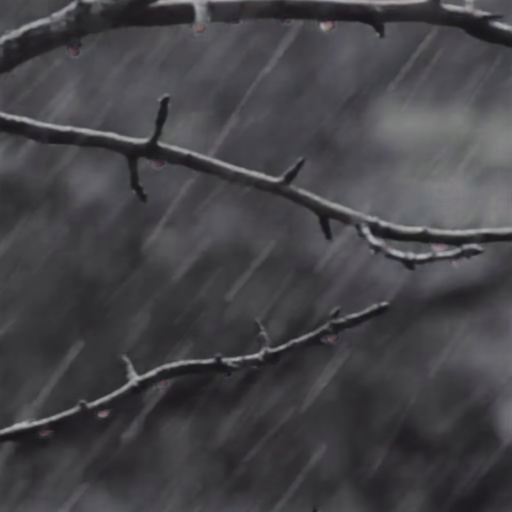} &
        \includegraphics[width=0.1\textwidth]{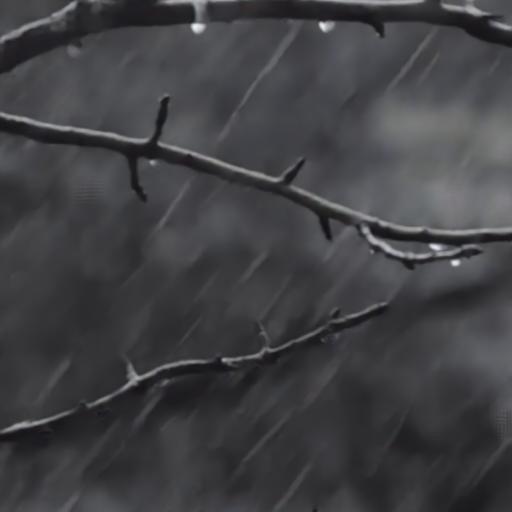} &
        \includegraphics[width=0.1\textwidth]{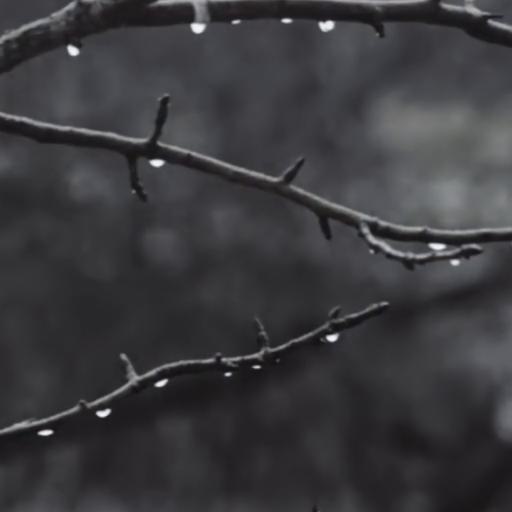} \\

        \includegraphics[width=0.1\textwidth]{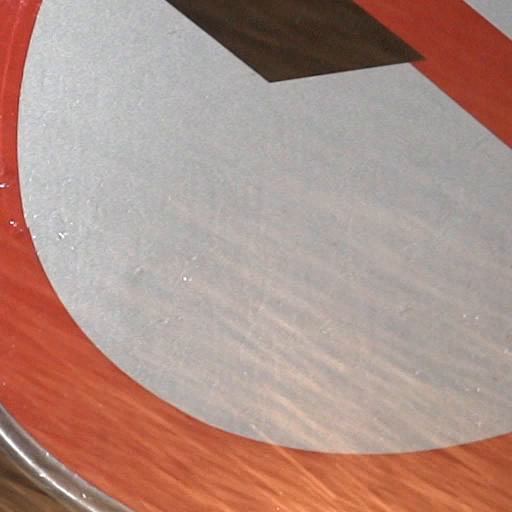} &
        \includegraphics[width=0.1\textwidth]{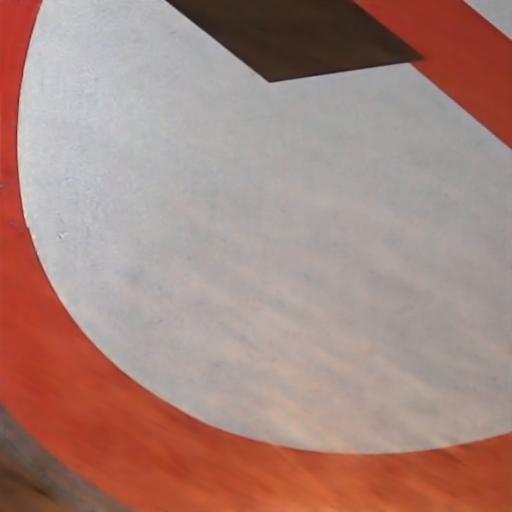} & 
        \includegraphics[width=0.1\textwidth]{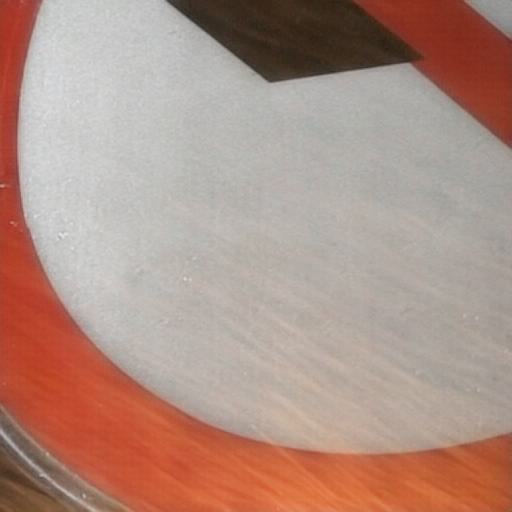} & 
        \includegraphics[width=0.1\textwidth]{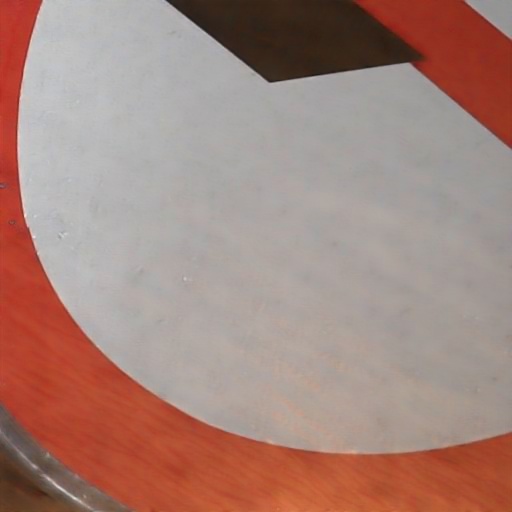} & 
        \includegraphics[width=0.1\textwidth]{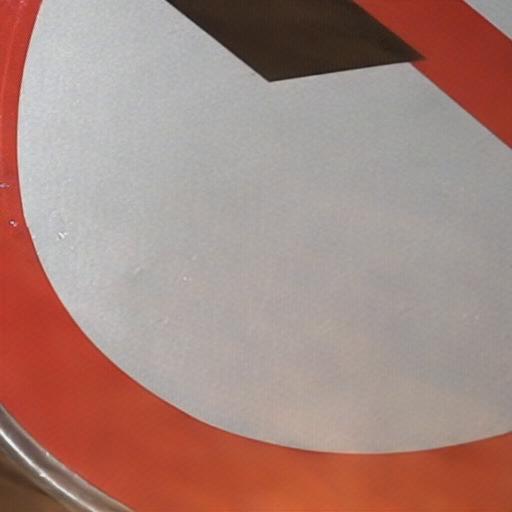} &
        \includegraphics[width=0.1\textwidth]{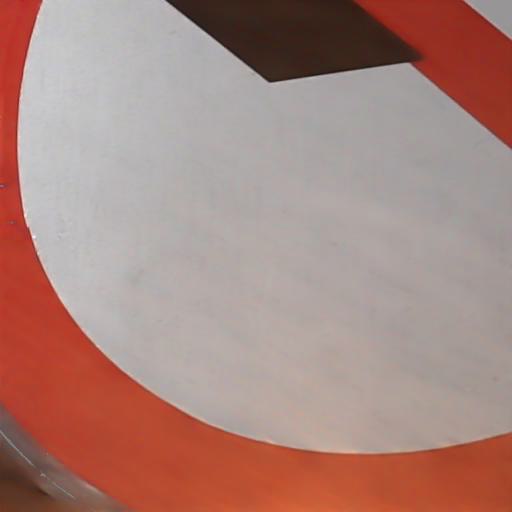} & 
        \includegraphics[width=0.1\textwidth]{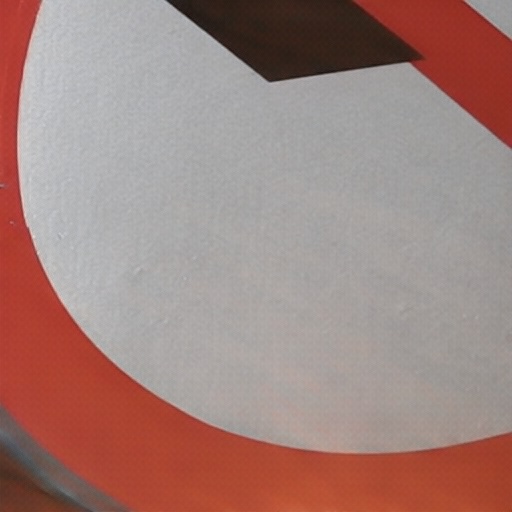} &
        \includegraphics[width=0.1\textwidth]{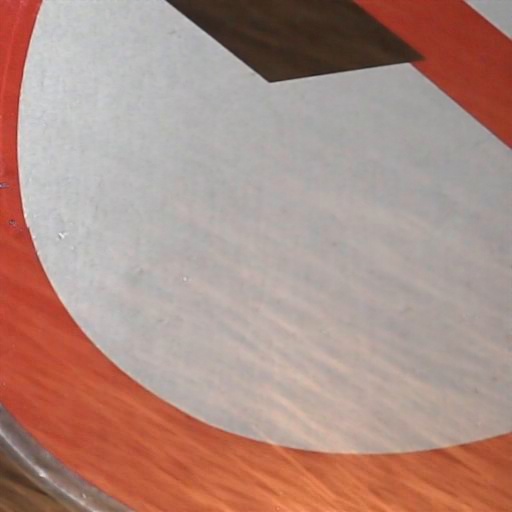} &
        \includegraphics[width=0.1\textwidth]{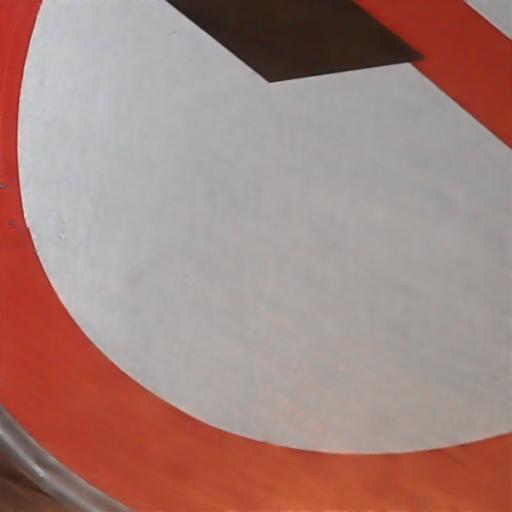} &
        \includegraphics[width=0.1\textwidth]{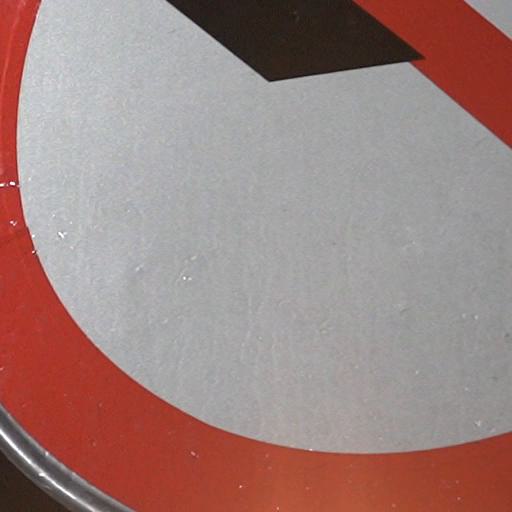} \\
        
        \includegraphics[width=0.1\textwidth]{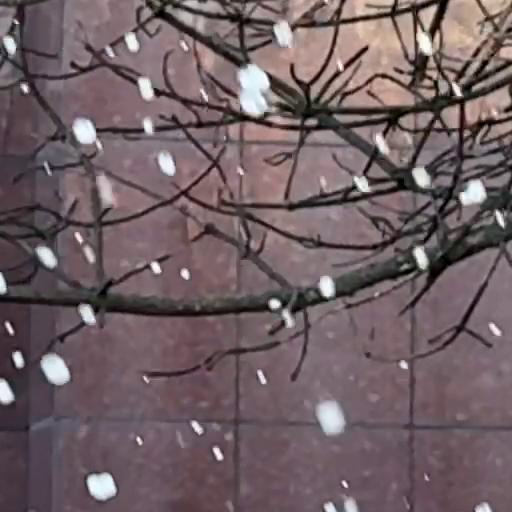} &
        \includegraphics[width=0.1\textwidth]{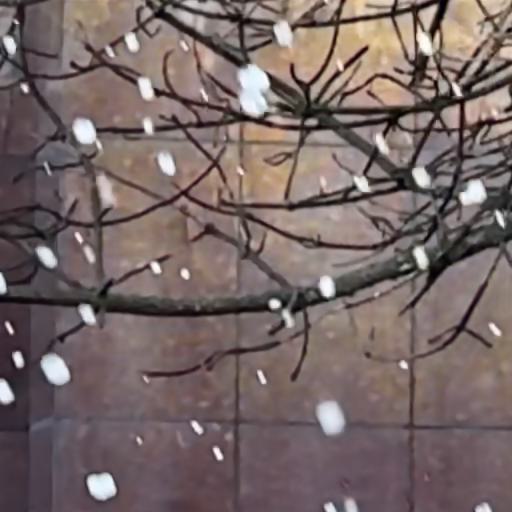} & 
        \includegraphics[width=0.1\textwidth]{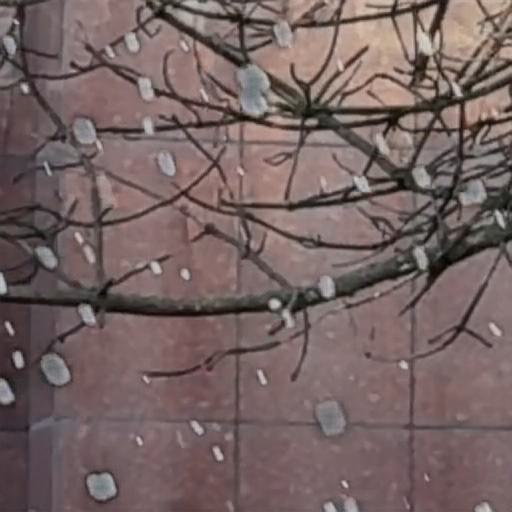} & 
        \includegraphics[width=0.1\textwidth]{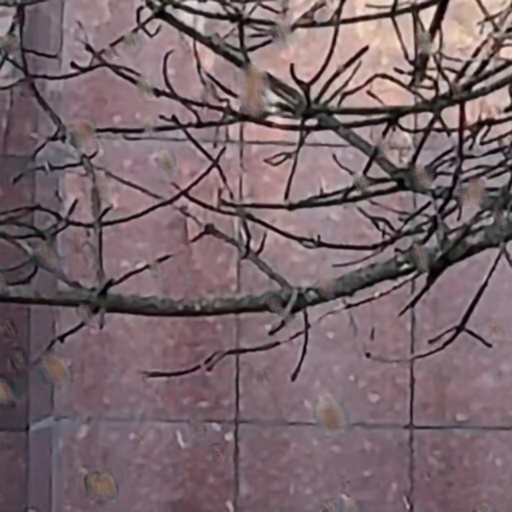} & 
        \includegraphics[width=0.1\textwidth]{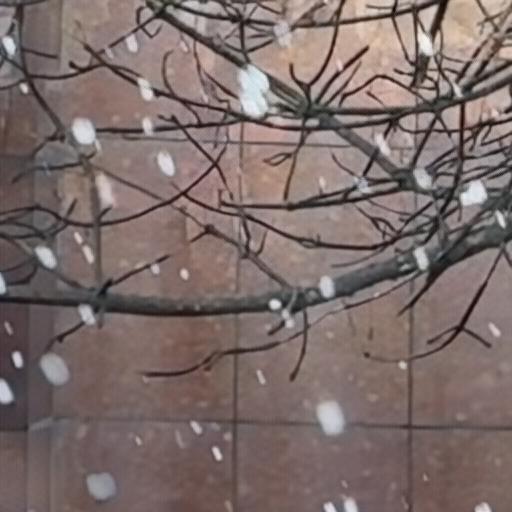} &
        \includegraphics[width=0.1\textwidth]{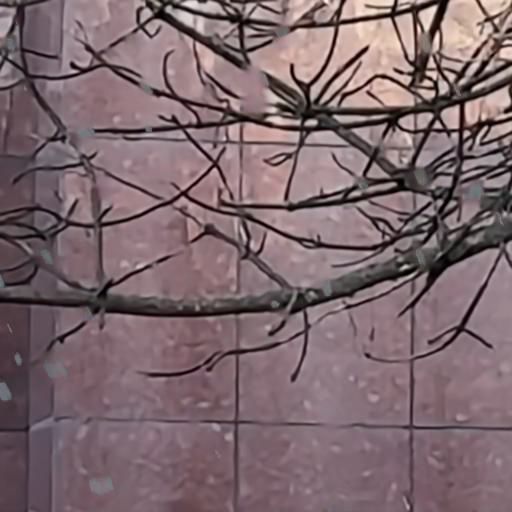} & 
        \includegraphics[width=0.1\textwidth]{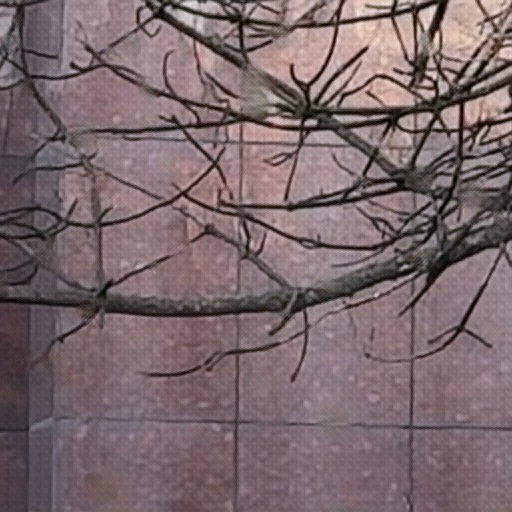} &
        \includegraphics[width=0.1\textwidth]{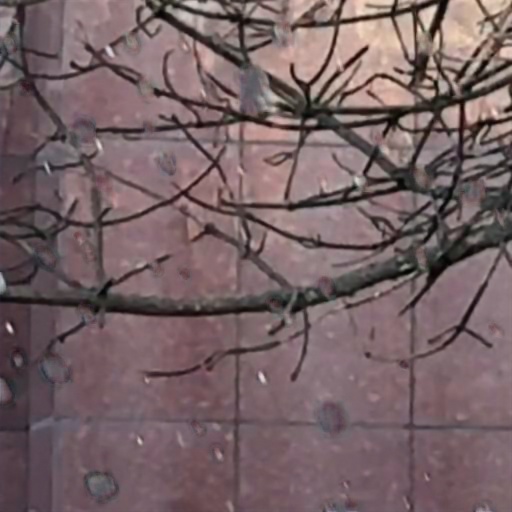} &
        \includegraphics[width=0.1\textwidth]{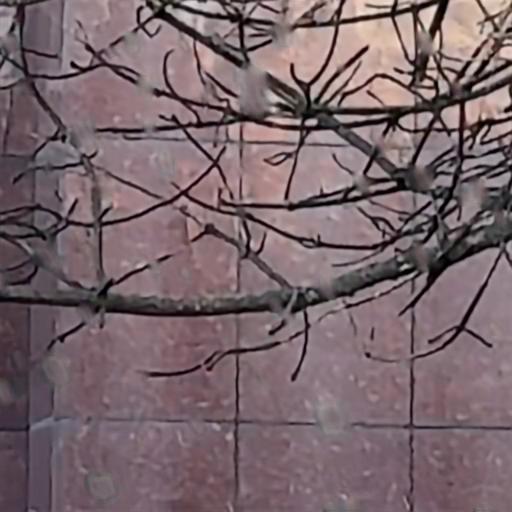} &
        \includegraphics[width=0.1\textwidth]{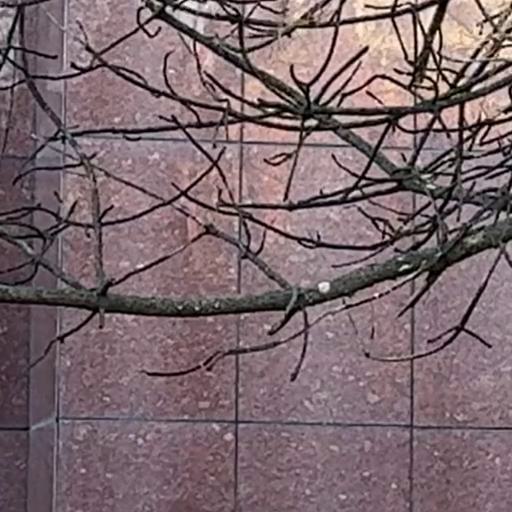} \\
        
        \includegraphics[width=0.1\textwidth]{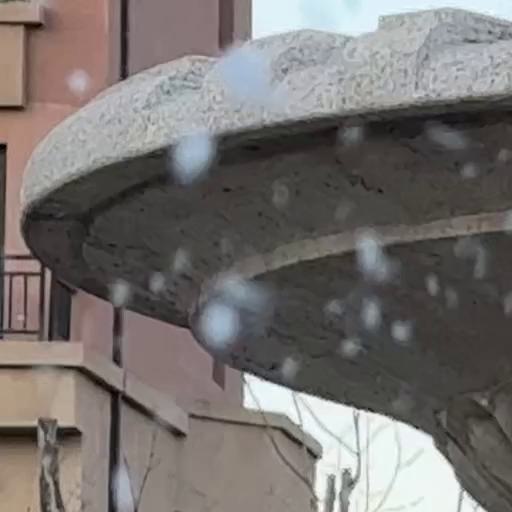} &
        \includegraphics[width=0.1\textwidth]{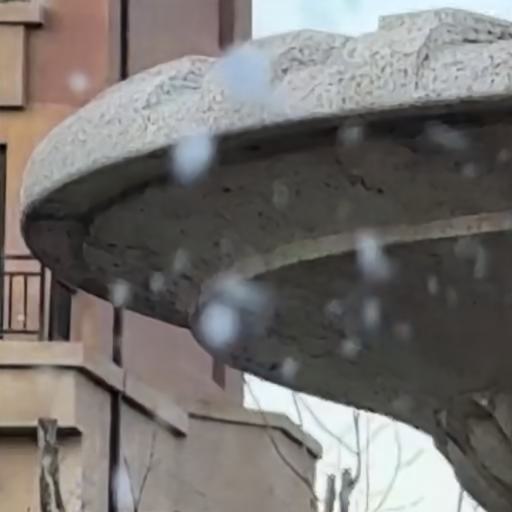} & 
        \includegraphics[width=0.1\textwidth]{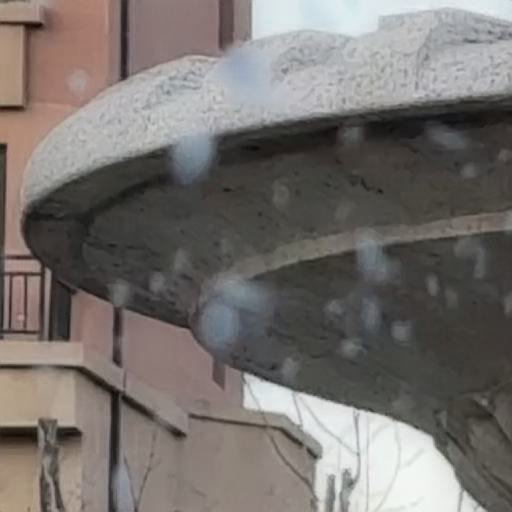} & 
        \includegraphics[width=0.1\textwidth]{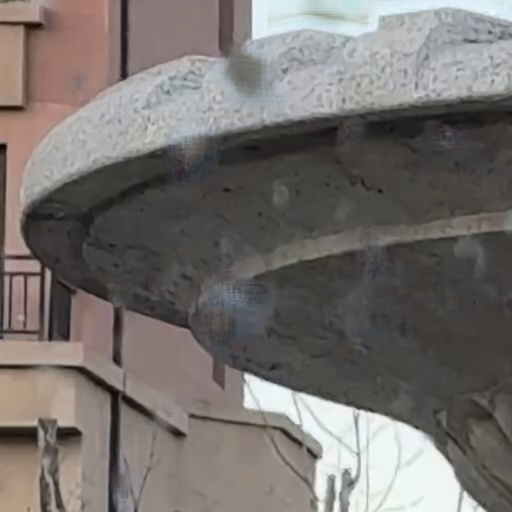} & 
        \includegraphics[width=0.1\textwidth]{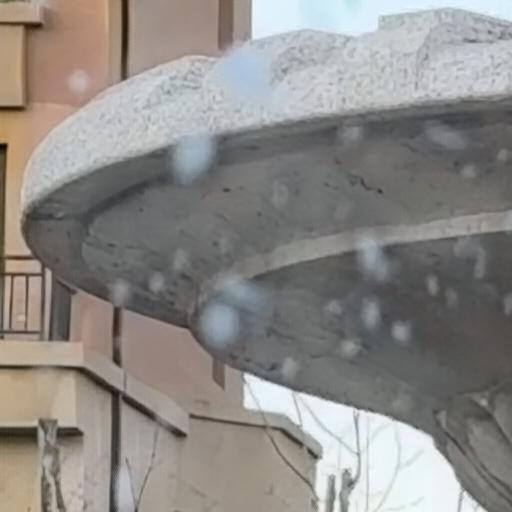} &
        \includegraphics[width=0.1\textwidth]{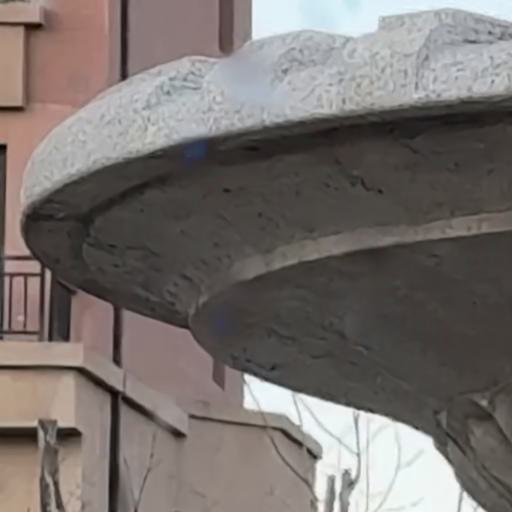} & 
        \includegraphics[width=0.1\textwidth]{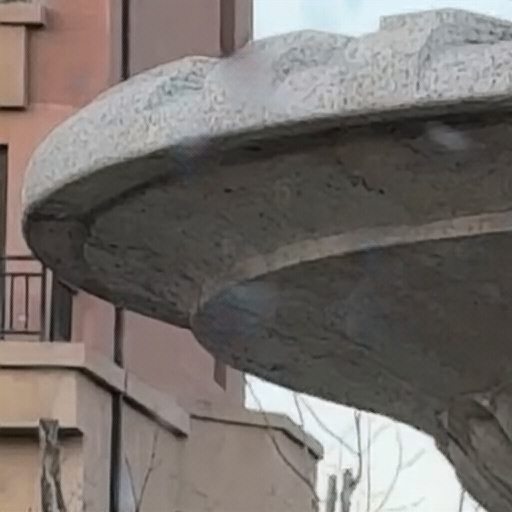} &
        \includegraphics[width=0.1\textwidth]{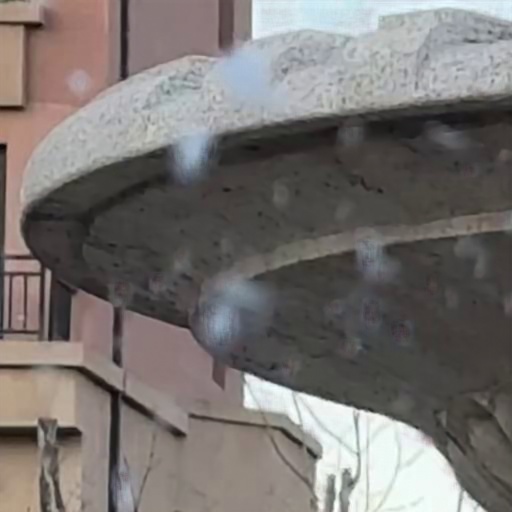} &
        \includegraphics[width=0.1\textwidth]{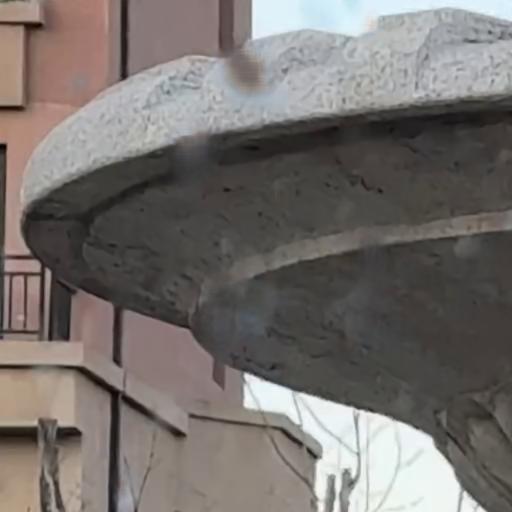} &
        \includegraphics[width=0.1\textwidth]{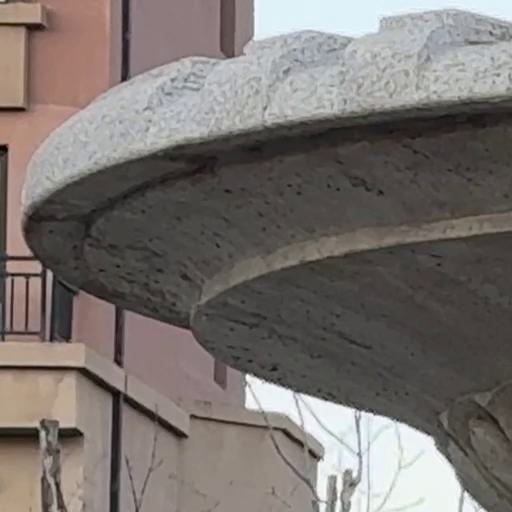} \\

        \includegraphics[width=0.1\textwidth]{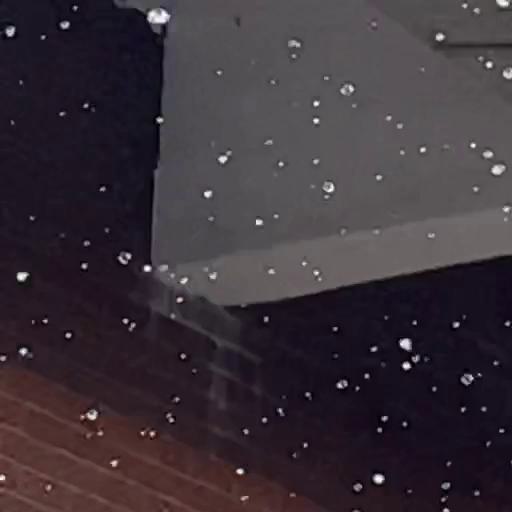} &
        \includegraphics[width=0.1\textwidth]{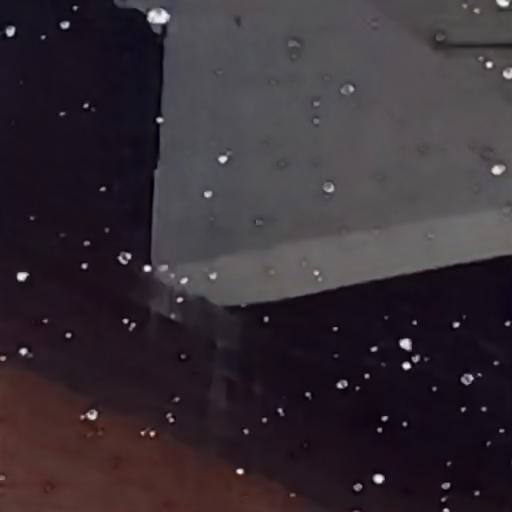} & 
        \includegraphics[width=0.1\textwidth]{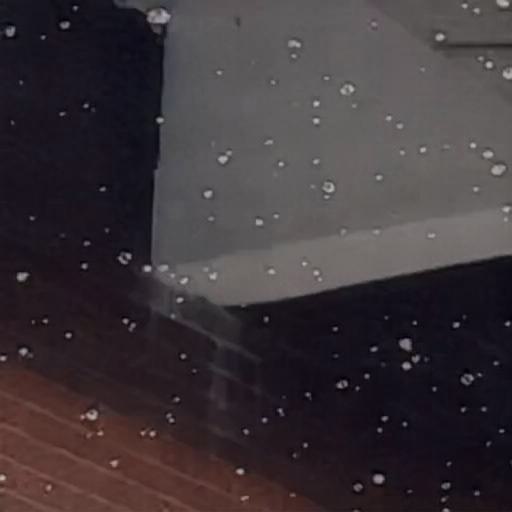} & 
        \includegraphics[width=0.1\textwidth]{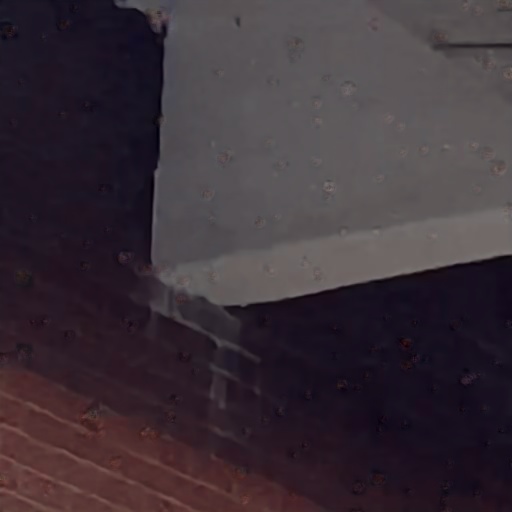} & 
        \includegraphics[width=0.1\textwidth]{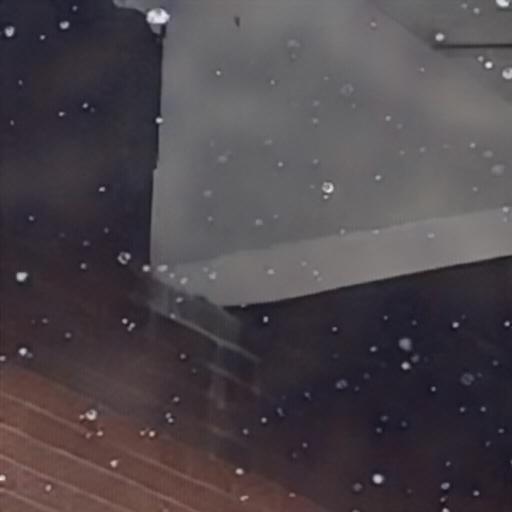} &
        \includegraphics[width=0.1\textwidth]{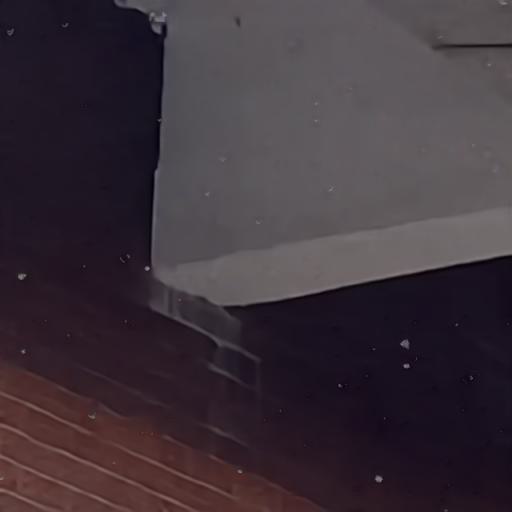} & 
        \includegraphics[width=0.1\textwidth]{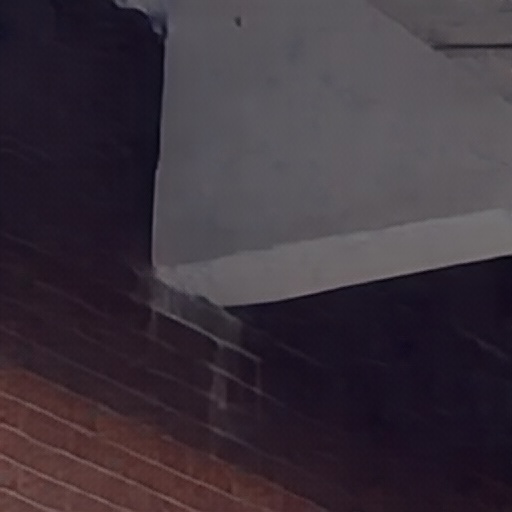} &
        \includegraphics[width=0.1\textwidth]{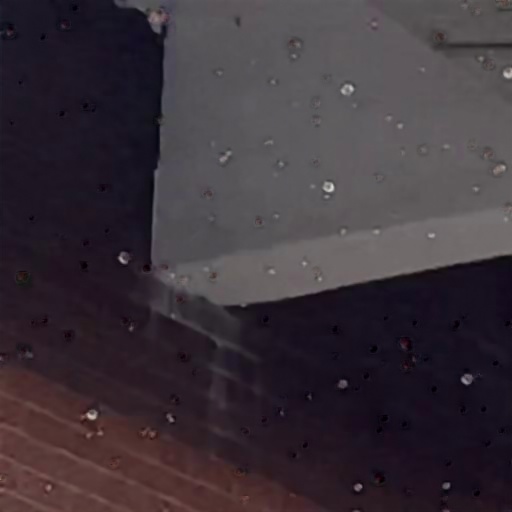} &
        \includegraphics[width=0.1\textwidth]{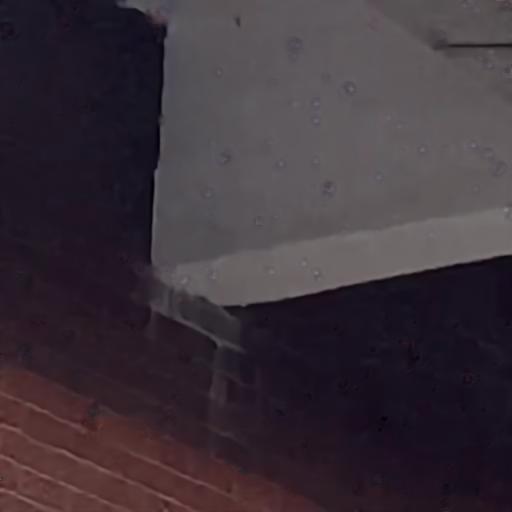} &
        \includegraphics[width=0.1\textwidth]{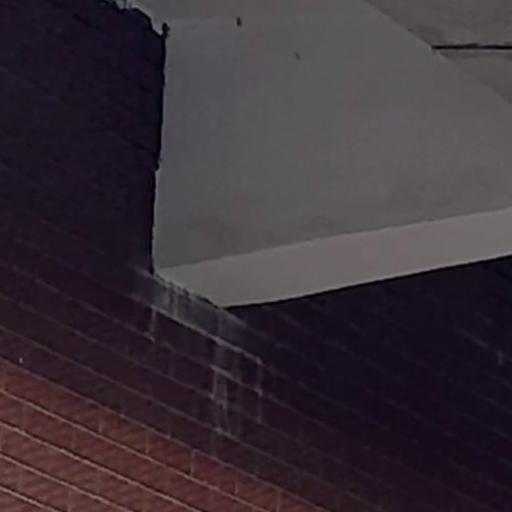} \\
        
        \makebox[0.1\textwidth][c]{Input} &
        \makebox[0.1\textwidth][c]{AirNet} &
        \makebox[0.1\textwidth][c]{TransWeather} &
        \makebox[0.1\textwidth][c]{PromptIR} &
        \makebox[0.1\textwidth][c]{WGWS-Net} &
        \makebox[0.1\textwidth][c]{DiffUIR} &
        \makebox[0.1\textwidth][c]{MWFormer} &
        \makebox[0.1\textwidth][c]{Histoformer} &
        \makebox[0.1\textwidth][c]{AdaIR} &
        \makebox[0.1\textwidth][c]{GT}
    \end{tabular}
    \caption{Qualitative results comparison of all-in-one methods.~We present visual results under three types of degradation—haze (rows 1–3), rain (rows 4–6), and snow (rows 7–9)—across both daytime and nighttime scenarios.}
    \label{fig:all-in-one}
\end{figure*}

{\flushleft\textbf{LQ capture}.}~We conduct data collection under various lighting conditions, including both daytime and nighttime scenes, to enhance the diversity of illumination in our dataset.
The LQ image acquisition process follows a consistent protocol in different types of weather.
For the \textbf{rain stage}, we used a rain generator equipped with interchangeable nozzles to simulate rainfall.
This setup allows for adjustable rain intensity, ranging from light to heavy rain, to better reflect real-world precipitation scenarios.
To further increase the diversity of the rain streak patterns, we sprayed water from different angles.
In nighttime scenes, we positioned the camera near light sources to capture more pronounced rain streaks, leveraging the increased visibility of rain under direct illumination.
For the \textbf{snow stage}, we added snow oil to a snow generator to produce realistic snowflakes.
By adjusting the machine’s pressure and spray direction, we generated a variety of snow patterns such as fine snow, drifting snow, and large snowflakes, thereby enriching the diversity of snowy degradations.
For the \textbf{haze stage}, we mixed water with haze oil in different ratios (e.g., 1:1, 1:1.5, or 1:1.7) and fed the mixture into a haze generator to produce haze with varying densities.
To improve the efficiency of gas–liquid conversion, the mixture was preheated for 1 to 2 minutes after powering on the generator.
Given that haze requires time to diffuse evenly across the scene, we included a delay between the haze emission and image capture to ensure a uniform haze distribution throughout the frame.

{\flushleft\textbf{GT capture}.}~For natural scenes, maintaining the spatial positions of scene components unchanged during recording poses significant challenges.
To this end, we need to strictly control the external environmental factors and internal technical parameters that affect the picture.
Specifically, during the shooting process, we prioritize capturing images under controlled lighting conditions and within a wind speed range of $0.1 m/s$ to $0.7 m/s$, in order to minimize brightness fluctuations and prevent unnecessary movement of lightweight objects such as trees, flowers, and leaves.
The inherent technical parameters encompass the uniformity of imaging devices, standardization of shooting parameters, consistency in camera viewpoints, as well as the elimination of image jitter caused by machine-generated airflow interference. 
In addition to the aforementioned control measures, to accurately acquire scene baseline data, we implement a four-stage GT image acquisition workflow.
Initial baseline recording of the pristine scene is conducted prior to the first water rain spraying.
Following complete evaporation of rain and thorough removal of residual water stains from the lens surface, secondary data collection is performed. 
Subsequent to snow spraying and adequate physical settling period, tertiary imaging is executed after meticulous elimination of condensed water from the optical assembly.
Finally, following haze dissipation, GT image acquisition should be performed after conducting comprehensive decontamination of condensate on the lens surface.

{\flushleft\textbf{GT image matching}.}~Although we take careful measures during data collection to minimize capture errors, unpredictable spatial and temporal misalignments may still occur between the GT and LQ image sequences.
To address this issue, we introduce a GT image matching strategy.
As illustrated in Figure~\ref{fig:data_pipeine}, we first crop a set of candidate patches from the captured GT sequence, prioritizing static objects such as buildings and parked vehicles, while excluding unstable regions like trees or vegetation.
Then, we select the best-matching patch based on appearance similarity to obtain a more spatially and temporally aligned GT reference.
Finally, we crop the corresponding region in the LQ image according to the GT patch location to construct an accurately aligned LQ-GT image pair.
This strategy effectively mitigates the impact of object motion or scene changes during capture and improves the consistency and quality of the paired data.

\subsection{Dataset Statistics and Comparisons}
Using the above pipelines, we capture a total of 50,000 pairs of degraded and corresponding clean images, from which we carefully select 42,002 high-quality image pairs to construct our dataset. 
To visualize the quantity distribution of different degradation types, a sunburst chart is illustrated in Figure~\ref{fig:data_show} (b).
The overall training and testing set of our proposed dataset contains 41,402 and 600 paired images, respectively.
The resolution of all images is $512 \times 512$. See Figure~\ref{fig:data_show} (c) for several examples.
The advantages of our benchmark offer two notable advantages over existing weather degradation datasets.
(i) {\bfseries real-world scenarios with larger-scale}: Our dataset comprises authentic scene image pairs captured across various outdoor scenarios and shooting periods, covering complex weather conditions (\eg, rain, snow, and haze) as well as diverse illumination settings (daytime and nighttime).
(ii) {\bfseries unified all-in-one weather benchmark}: Existing methods combine several small-scale synthetic datasets as training data for all-in-one image restoration (as shown in Table~\ref{tab:dataset_survey}), limiting real-world applications and all-in-one model development. Our dataset aims to provide a unified platform for training and evaluating all-in-one weather restoration models, advancing the field toward more general and practical solutions.

\section{Benchmark Evaluation}

\subsection{Evaluation Settings}

{\flushleft\textbf{Comparison Baselines}.}~To verify the effectiveness and applicability of the proposed dataset, we conduct extensive experiments on a variety of image restoration methods. These methods are divided into three categories: task-specific methods~(DehazeFormer~\cite{song2023vision}, DCMPNet~\cite{zhang2024depth},
 DRSformer~\cite{chen2023learning}, NeRD-Rain~\cite{chen2024bidirectional}, and SnowForme-\\r~\cite{chen2022snowformer}), task-general methods~(MPRNet~\cite{zamir2021multi} and Restormer~\cite{zamir2022restormer}), and all-in-one methods~(AirNet~\cite{li2022all}, TransWeather~\cite{valanarasu2022transweather}, PromptIR~\cite{potlapalli2023promptir}, WGWS-Net~\cite{zhu2023learning}, DiffUIR~\cite{zheng2024selective}, MWFormer~\cite{zhu2024mwformer}, Histoformer~\cite{sun2024restoring}, and AdaIR~\cite{cui2025adair}).

{\flushleft\textbf{Implementation details and metrics}.}~For a fair comparison, we uniformly set the patch size to 128 for all selected methods, while keeping the other parameters consistent with their original settings.
All experiments are conducted on a single NVIDIA RTX 4090 GPU.
We adopt PSNR, SSIM, and LPIPS computed on the Y channel as evaluation metrics.

\subsection{Quantitative Comparisons}
We conduct a comprehensive comparison across three typical weather degradation tasks: dehazing, deraining, and desnowing.
As shown in Table~\ref{tab:methods}, although task-specific methods (\eg, DehazeFormer~\cite{song2023vision}, DRSformer~\cite{chen2023learning}, SnowFormer~\cite{chen2022snowformer}) are tailored for individual degradation types, their performance is not consistently superior within their designated tasks.
Surprisingly, some All-in-One methods such as MWFormer~\cite{zhu2024mwformer} and AdaIR~\cite{cui2025adair} achieve competitive or even better results in single-task scenarios.
For instance, MWFormer~\cite{zhu2024mwformer} outperforms DehazeFormer~\cite{song2023vision} in the dehazing task with a PSNR of 24.42dB versus 24.12dB, while also maintaining robust performance across the other tasks.
On average, all-in-One methods exhibit stronger generalization ability and more balanced restoration quality across diverse weather conditions.
This highlights the necessity and effectiveness of building unified models that can handle various weather degradations within a single framework, making them more practical for real-world applications.

\subsection{Qualitative Comparisons}
As illustrated in Figure~\ref{fig:all-in-one}, some methods, such as AdaIR~\cite{cui2025adair}, consistently produce visually pleasing and artifact-free results in all three scenarios, demonstrating strong adaptability and robustness to various weather degradations.
In contrast, other methods exhibit uneven performance across tasks.
For example, PromptIR~\cite{potlapalli2023promptir} delivers visually sharp and clean results in the desnowing task, but fails to effectively remove haze, often leaving residual haze or over-enhancing the scene.
WGWS-Net~\cite{zhu2023learning} performs well in the deraining task, but its performance in the dehazing and desnowing tasks is unsatisfactory.

\begin{table}[!t]
\footnotesize
\renewcommand{\arraystretch}{1.3}
\setlength{\tabcolsep}{1.6pt}
\caption{Quantitative comparison of training on our dataset and the hybrid dataset with testing on the real-world dataset.~Where `*' indicates models trained on our WeatherBench dataset.~Best results are highlighted in \textbf{bold}.}
\begin{tabular}{c|cc|cc|cc}
\hline
Methods & PromptIR*~\cite{potlapalli2023promptir} & PromptIR & DiffUIR*~\cite{zheng2024selective} & DiffUIR & AdaIR*~\cite{cui2025adair}  & AdaIR   \\ \hline
PSNR$\uparrow$    & \textbf{20.4513}   & 20.1860  & \textbf{20.3039}  & 19.9420 & \textbf{18.8749} & 18.2891 \\
SSIM$\uparrow$    & 0.6510    & \textbf{0.6840}   & \textbf{0.6730}   & 0.6726  & \textbf{0.5455}  & 0.5412  \\ \hline
\end{tabular}
\label{tab:train_on_diff}
\end{table}

\begin{figure}[!t]
  \centering
  \setlength{\tabcolsep}{1pt}
  \renewcommand{\arraystretch}{0.6}
  \begin{tabular}{cccc}
    \includegraphics[width=0.23\linewidth]{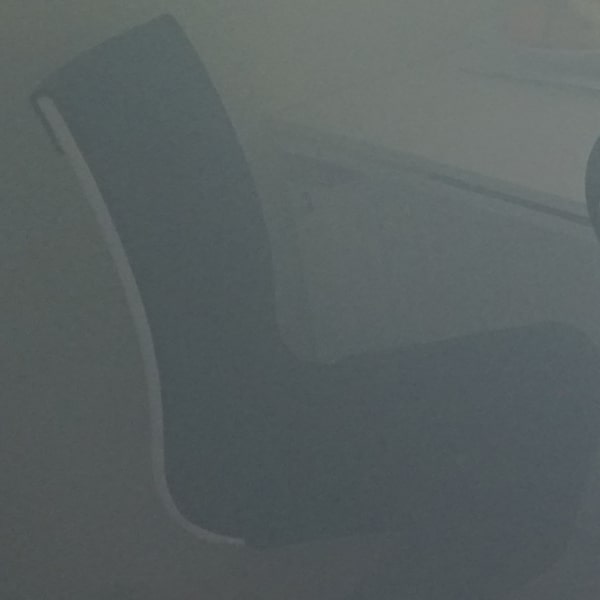} &
    \includegraphics[width=0.23\linewidth]{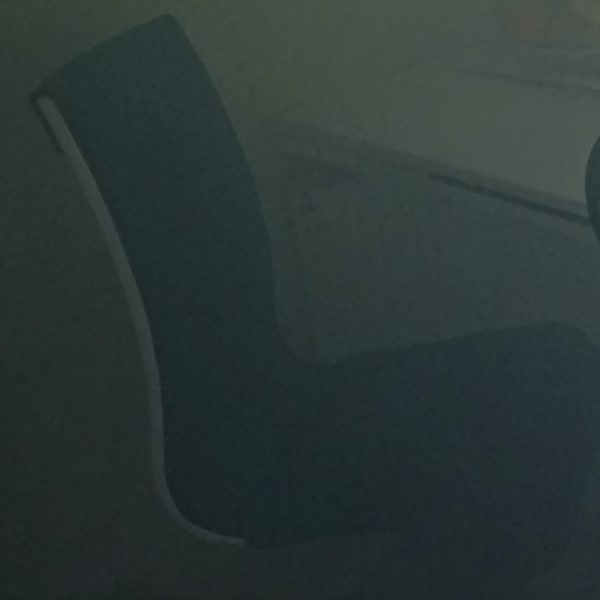} &
    \includegraphics[width=0.23\linewidth]{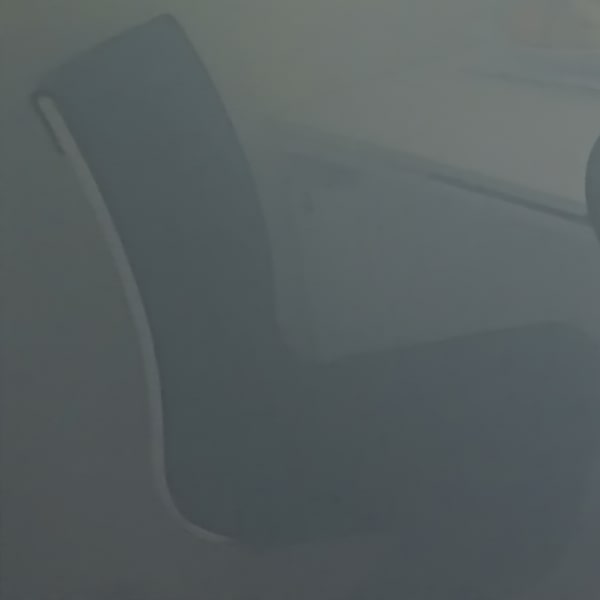} &
    \includegraphics[width=0.23\linewidth]{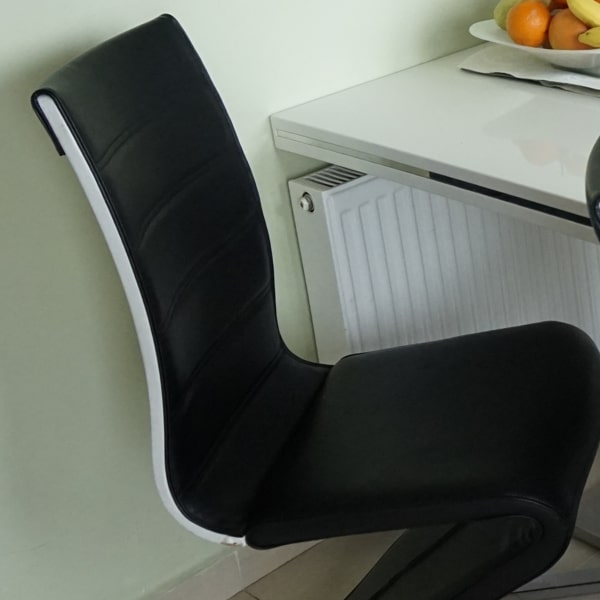} \\
    
    \includegraphics[width=0.23\linewidth]{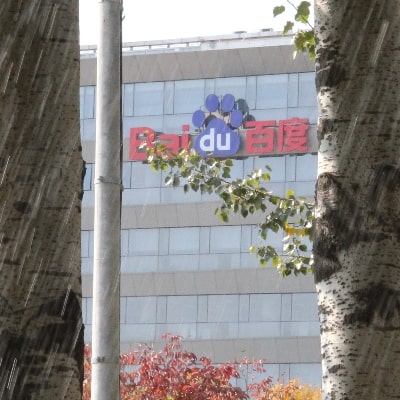} &
    \includegraphics[width=0.23\linewidth]{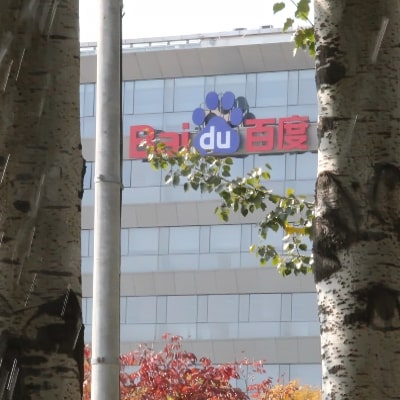} &
    \includegraphics[width=0.23\linewidth]{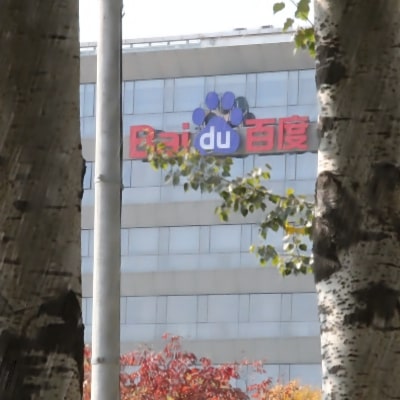} &
    \includegraphics[width=0.23\linewidth]{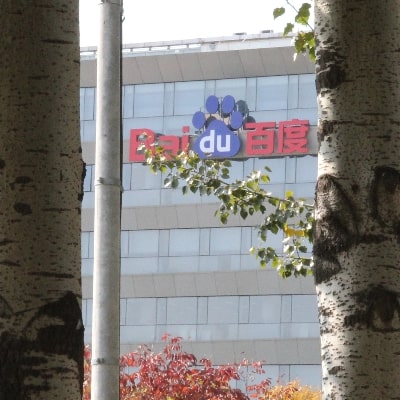} \\
    
    \includegraphics[width=0.23\linewidth]{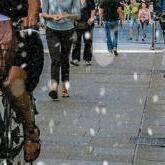} &
    \includegraphics[width=0.23\linewidth]{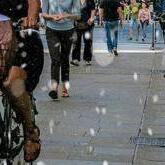} &
    \includegraphics[width=0.23\linewidth]{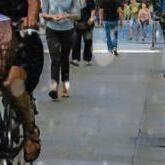} &
    \includegraphics[width=0.23\linewidth]{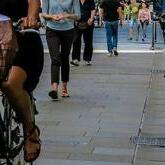} \\
    
    Input & AdaIR & AdaIR* & GT
  \end{tabular}
  \caption{Qualitative results of generalization ability on real-world data.~Where `*' indicates models trained on our WeatherBench dataset and achieving superior results.}
  \label{fig:train_on_other_data}
\end{figure}

\subsection{Generalization Ability on Real-World Data}
To verify the generalization ability of models trained on our proposed dataset to real-world scenarios, we construct a hybrid dataset composed of several commonly used datasets (\eg ~CSD~\cite{chen2021all}, Rain-\\13k~\cite{jiang2020multi}, RESIDE-$\beta$-OTS~\cite{li2018benchmarking}) as a comparison.
Multiple representative methods are trained on both the mixed dataset and our dataset, and evaluated on three real-world benchmarks: Dense-Haze~\cite{ancuti2019dense}, RainDS-real~\cite{quan2021removing}, and Snow100K~\cite{liu2018desnownet}.
As shown in Table~\ref{tab:train_on_diff}, models trained on our dataset generally achieve the best quantitative performance, demonstrating superior generalization ability to real-world data.
Qualitative results in Figure~\ref{fig:train_on_other_data} further confirm that models trained on our dataset produce visually better restoration results.

\section{Conclusion}
%
In this work, we present WeatherBench, a real-world all-in-one image restoration benchmark addressing the limitations of synthetic datasets. Unlike previous methods that combine disparate single-weather datasets, WeatherBench offers high-quality, aligned image pairs under real rain, snow, and haze across diverse scenes. Our all-in-one weather acquisition system enables supervised learning. Extensive experiments highlight the strengths and weaknesses of different approaches, underscoring the need for unified, robust restoration models in realistic conditions.

\begin{acks}
This work was supported in part by the Scientific Research Project of the Education Department of Liaoning Province (LJ212410152006).
\end{acks}

\bibliographystyle{ACM-Reference-Format}
\balance
\bibliography{sample-base}










\end{document}